\documentclass{article}

\usepackage{microtype}
\usepackage{graphicx}
\usepackage{subfigure}
\usepackage{booktabs} 
\usepackage{multirow}
\usepackage{xspace}

\usepackage{hyperref}
\usepackage{url}

\usepackage{amsmath}
\usepackage{amssymb}
\usepackage{mathtools}
\usepackage{amsthm}
\usepackage{makecell}
\usepackage{paralist}

\usepackage[accepted]{icml2023}

\usepackage[capitalize,noabbrev]{cleveref}

\usepackage[textsize=tiny]{todonotes}

\usepackage{amsmath,amsfonts,bm}

\def\eqref#1{equation~\ref{#1}}

\def\1{\bm{1}}

\def\eps{{\epsilon}}

\def\rmLambda{{\mathbf{\Lambda}}}
\def\rmTheta{{\mathbf{\Theta}}}
\def\rmA{{\mathbf{A}}}

\def\rmD{{\mathbf{D}}}

\def\rmH{{\mathbf{H}}}
\def\rmI{{\mathbf{I}}}

\def\rmL{{\mathbf{L}}}
\def\rmM{{\mathbf{M}}}

\def\rmS{{\mathbf{S}}}

\def\rmU{{\mathbf{U}}}

\def\rmW{{\mathbf{W}}}

\def\rmZ{{\mathbf{Z}}}

\def\vtheta{{\bm{\theta}}}
\def\vlambda{{\bm{\lambda}}}

\def\vf{{\bm{f}}}

\def\vt{{\bm{t}}}

\DeclareMathAlphabet{\mathsfit}{\encodingdefault}{\sfdefault}{m}{sl}
\SetMathAlphabet{\mathsfit}{bold}{\encodingdefault}{\sfdefault}{bx}{n}

\def\gD{{\mathcal{D}}}
\def\gE{{\mathcal{E}}}
\def\gF{{\mathcal{F}}}
\def\gG{{\mathcal{G}}}

\def\gM{{\mathcal{M}}}

\def\gV{{\mathcal{V}}}

\def\sR{{\mathbb{R}}}

\theoremstyle{plain}
\newtheorem{theorem}{Theorem}[section]
\newtheorem{proposition}[theorem]{Proposition}

\theoremstyle{definition}
\newtheorem{definition}[theorem]{Definition}

\theoremstyle{remark}

\newcommand{\method}{\textsc{PDF}\xspace}

\begin{document}
	
	\twocolumn[
	\icmltitle{Towards Better Graph Representation Learning with Parameterized Decomposition \& Filtering}
	
	
	

	\begin{icmlauthorlist}
		\icmlauthor{Mingqi Yang}{dlut}
		\icmlauthor{Wenjie Feng}{ids}
		\icmlauthor{Yanming Shen}{dlut}
		\icmlauthor{Bryan Hooi}{ids,soc}
	\end{icmlauthorlist}
	
	\icmlaffiliation{dlut}{Dalian University of Technology, China}
	\icmlaffiliation{ids}{Institute of Data Science, National University of Singapore, Singapore}
	\icmlaffiliation{soc}{School of Computing, National University of Singapore, Singapore}
	
	\icmlcorrespondingauthor{Yanming Shen}{shen@dlut.edu.cn}
	
	\icmlkeywords{Machine Learning, ICML}

	\vskip 0.3in
	]
	
	
	
	\printAffiliationsAndNotice{}  

	\begin{abstract}
		Proposing an effective and flexible matrix to represent a graph is a fundamental challenge that has been explored from multiple perspectives, e.g., filtering in Graph Fourier Transforms. In this work, we develop a novel and general framework which unifies many existing GNN models from the view of parameterized decomposition and filtering, and show how it helps to enhance the flexibility of GNNs while alleviating the smoothness and amplification issues of existing models. Essentially, we show that the extensively studied spectral graph convolutions with learnable polynomial filters are constrained variants of this formulation, and releasing these constraints enables our model to express the desired decomposition and filtering simultaneously. Based on this generalized framework, we develop models that are simple in implementation but achieve significant improvements and computational efficiency on a variety of graph learning tasks. Code is available at \url{https://github.com/qslim/PDF}.
	\end{abstract}

	\section{Introduction}
	
	Graph Neural Networks (GNNs) have emerged as a powerful and promising technique for 
	representation learning on graphs and have been widely applied to various applications.
	A large number of GNN models have been proposed, 
	including spectral graph convolutions, spatial message-passing, and even Graph Transformers, 
	for boosting performance or resolving existing defects, 
	e.g., the oversmoothing issue~\citep{li2018deeper,oono2020graph,huang2020tackling} or expressive power~\citep{xu2018how,morris2019weisfeiler,sato2020survey}.
	This raises the natural question of: \emph{how are these different models related to one another?}
	Consequently, when conducting learning tasks on a graph with these models, 
	the fundamental questions we aim to address are:
	how to \emph{construct an effective adaptive matrix representation} for the graph topology, 
	and how to \emph{flexibly capture the interactions between multichannel graph signals}.
	
	Under the paradigm of spectral graph convolution theories,
	the Laplacian and its variants are used as the matrix representation of the graph to 
	maintain theoretical consistency~\citep{chung1997spectral,hammond2011wavelets,shuman2013emerging,defferrard2016convolutional}.
	This induces spectral GNNs, which have become a popular class of GNNs with performance guarantees, 
	including GCN~\citep{kipf2017semi}, SGC~\citep{pmlr-v97-wu19e}, S$^2$GC~\cite{zhu2020simple}, and others~\citep{klicpera_predict_2019,klicpera2019diffusion,chenWHDL2020gcnii}.
	All of them utilize the eigenvectors of the (normalized) Laplacian 
	for Graph Fourier Transform and design different graph signal filters in the frequency domain.
	
	Beyond spectral models, spatial GNNs have gradually become prominent in the research community, 
	which consider graph convolution within the message-passing framework~\citep{gilmer2017neural,xu2018how,corso2020pna,yang2020breaking}, 
	and the advantage of flexible implementation makes them favorable for the graph-level prediction task.
	Interestingly, the aggregation design, which plays a central role in the spatial GNNs, 
	also implicitly corresponds to a specific matrix representation for the graph.
	
	Recently, transformers have shown promising performance on molecular property predictions~\citep{ying2021transformers,kreuzer2021rethinking,bastos2022how,min2022transformer,rampasek2022GPS}.
	These models apply transformers with positional encoding, structural encoding, 
	and other techniques to graph data, while they actually also refer to specific matrix representations.
	Besides, several studies in the field of graph signal processing also show that 
	the matrix representation can be flexible as long as it reflects the topology of 
	the underlying graph~\citep{dong2016learning,deri2017spectral,ortega2018graph}.
	Graph Shift Operator (GSO) proposes a group of feasible matrix representations~\citep{sandryhaila2013discrete}, 
	and allows GNNs to obtain better performance for graph learning tasks by 
	learning matrix representations from a group of GSOs~\citep{dasoulas2021learning}.
	
	Accordingly, designing a flexible and suitable matrix representation for
	the underlying graph plays an intrinsic role in improving GNNs on various tasks.
	Although many specific choices exist, each of them has its own limitations and 
	no single matrix representation is suitable for every task~\citep{butler2017spectral}.
	Therefore, it is highly important to systematically explore the differences among these different models and 
	automatically find the most suitable matrix representation, 
	while capturing the complex interaction of multichannel signals for a graph learning task.
	
	In this paper, we show how various graph matrix representations applied to graph learning can be interpreted as different (Decomposition, Filtering) operations over input signals assigned to the graph.
	Then, based on this understanding, we propose to learn (Decomposition, Filtering) as a whole, fundamentally different from existing spectral graph convolutions with learnable filters.
	Correspondingly, the objective is extended from learning a suitable filter to learning a suitable graph matrix representation, i.e. (Decomposition, Filtering).
	
	To achieve this, we propose a novel  Parameterized-$(\gD, \gF)$ framework which aims to learn a suitable (Decomposition, Filtering) for input signals.
	It inspires the use of more expressive learnable mappings on graph topology to enlarge the learning space of graph matrix representations.
	Also, Parameterized-$(\gD, \gF)$ serves as a general framework that unifies existing GNNs ranging from the original spectral graph convolutions to the latest Graph Transformers.
	For example, spectral graph convolution corresponds to fixed $\gD$ and parameterized $\gF$ via the lens of Parameterized-$(\gD, \gF)$, and recent Graph Transformers, which improve performance by leveraging positional/structural encodings, potentially result in more effective $(\gD, \gF)$ on input signals.
	Parameterized-$(\gD, \gF)$ also inspires new insight into the widely-studied smoothing and amplification issues in GNNs, which serves as the motivation of our proposed solution as well.
	Our main contributions include, 
	\begin{compactenum}
		\item We present Parameterized-$(\gD, \gF)$, which addresses the deficiencies of learning filters alone, and also reveals the connections between various existing GNNs;
		\item With Parameterized-$(\gD, \gF)$ framework, we develop a model with a simple implementation that achieves superior performance while preserving computational efficiency.
	\end{compactenum}

	\section{Preliminaries}
	\label{sec:prelim}
	
	Consider an undirected graph $G = (\gV, \gE)$ with vertex set $\gV$ and edge set $\gE$.
	Let $\rmA \in \mathbb{R}^{n \times n}$ be the adjacency matrix ($\rmW \in \mathbb{R}^{n \times n}$ if $G$ is weighted)
	with corresponding degree matrix $\rmD = \mathrm{diag}(\rmA \bm{1}_n)$, $\rmL = \rmD - \rmA$ be the Laplacian, 
	$\tilde \rmA=\rmA+\rmI$ and $\tilde \rmD=\rmD+\rmI$.
	Let $\vf \in \mathbb{R}^{n}$ be the single-channel graph signal assigned on $G$, 
	and $\rmH \in \mathbb{R}^{n \times d}$ be the $d$-channel graph signal or node feature matrix with $d$ dimensions.
	We use $[K]$ to denote the set $\{0, 1, 2,\dots, K\}$.
	
	\textbf{Graph Representation.}
	A graph topology can be expressed with various matrix representations, 
	e.g., (normalized) adjacency, Laplacian, GSO~\citep{sandryhaila2013discrete}, etc.
	\cite{dong2016learning,deri2017spectral,ortega2018graph} show that the representation matrix used in graph signal processing 
	can be flexible as long as it reflects the graph topology, and different representations lead to different signal models.
	In recent GNN studies, the involved graph representations are even more flexible~\citep{ying2021transformers}.
	Since they are all induced from the same graph topology, we use the graph representation space $\gM_{G}$ 
	to denote the set of all possible representations for a graph $G$.
	A summary of existing strategies for building a graph representation $\rmS \in \gM_{G}$ is provided in Appendix~\ref{appe:graph_representation}.
	In this work, we only consider undirected graphs with $\gM_{G}$ only involving symmetric matrices.
	
	\section{Parameterized-$(\gD, \gF)$}
	\label{sec:sepc_gnn}
	
	\subsection{Generalizing Spectral Graph Convolution}
	
	Given a graph $G$ with the Laplacian $\rmL = \rmU \rmLambda \rmU^{\top}$, $\rmLambda = \mathrm{diag}(\vlambda)$, 
	a filter $\vt$ and a graph signal $\vf$ assigned on $G$, the spectral graph convolution of $\vf$ with $\vt$ on $G$ 
	leveraging convolution theorem~\citep{hammond2011wavelets,defferrard2016convolutional} is defined as
	\begin{equation}
		\label{equ:gc}
		\begin{aligned}
			\vf^{\prime} &= \vt *_{\rmL} \vf = \rmU ((\rmU^{\top} \vt) \odot (\rmU^{\top} \vf)) \\
			&\approx \rmU (g_{\vtheta}(\vlambda) \odot (\rmU^{\top} \vf)) = \rmU (g_{\vtheta}(\rmLambda) (\rmU^{\top} \vf)) \\
			&= g_{\vtheta}(\rmL) \vf,
		\end{aligned}
	\end{equation}
	where $\hat{\vf} = \rmU^{\top} \vf$ is the Graph Fourier Transform,
	and $\vf = \rmU \hat{\vf}$ is its inverse transform over the graph domain, 
	$g_{\vtheta}(\vlambda)$ is the polynomial over $\vlambda$ and is used to 
	approximate the transformed filter $\hat{\vt} = \rmU^{\top} \vt$.
	Although various graph convolutions have been proposed, they are all under the formulation in Eq.~\ref{equ:gc}, 
	and most of them enhance $g_{\vtheta}(\vlambda)$ with sophisticated designed polynomials with learnable coefficients.
	
	As $\rmU$ is column orthogonal and $g_{\vtheta}(\rmLambda)$ is diagonal, 
	the convolution $\vt *_{\rmL} \vf = \rmU (g_{\vtheta}(\rmLambda) (\rmU^{\top} \vf))$ in Eq.~\ref{equ:gc} 
	can be divided into three individual steps:
	\begin{compactenum}[1)]
		\item $\hat{\vf} = \rmU^{\top} \vf$: \emph{decomposition} (or \emph{transformation});
		\item $\hat{\vf}^{\prime} = g_{\vtheta}(\rmLambda) \hat{\vf}$: \emph{filtering} (or \emph{scaling}) with $g_{\vtheta}(\rmLambda)$;
		\item $\vf^{\prime} = \rmU \hat{\vf}^{\prime}$: \emph{reverse transformation}.
	\end{compactenum}
	Consequently, we can interpret the convolution as above that filter $\vf$ with 
	$g_{\vtheta}(\rmLambda)$ under the decomposition $\rmU^{\top}$.
	Without loss of generality, we use $\gD$ to denote the involved decomposition and its reversion, 
	$\gF$ to denote the involved filtering, and then the above operation is represented as 
	$(\gD, \gF)_{g_{\vtheta}(\rmL)}(\vf)$ 
	which indicates the applied $(\gD, \gF)$ on $\vf$ is provided by $g_{\vtheta}(\rmL)$.
	The options of $(\gD, \gF)_{g_{\vtheta}(\rmL)}$ in general polynomial graph filters are only restricted within $\{g_{\vtheta}(\rmL)|\vtheta\in\mathbb R^k\}\subset\gM_{G}$ where $k$ is the order of the polynomial and $\vtheta$ is the polynomial coefficients.
	Note that any $\rmS \in \gM_{G}$ is a symmetric matrix with the unique eigendecomposition $\rmS= \rmU^{\prime} \hat{\rmLambda}^{\prime} \rmU^{\prime\top}$, where $\rmU^{\prime}$ and $\hat{\rmLambda}^{\prime}$ can serve as a feasible $(\gD, \gF)$. It is desirable to extend $(\gD, \gF)$ beyond $\{g_{\vtheta}(\rmL)|\vtheta\in\mathbb R^k\}$. Hence we have for any $\rmS \in \gM_{G}$,
	\begin{equation}
		\nonumber
		\begin{aligned}
			\rmS\vf &= \rmU^{\prime} \hat{\rmLambda}^{\prime} (\rmU^{\prime\top} \vf) = (\gD, \gF)_{\rmS}(\vf).
		\end{aligned}
	\end{equation}
	Still, $\rmS\vf$ refers to filtering $\vf$ with $\hat{\rmLambda}^{\prime}$ under the decomposition $\rmU^{\prime\top}$, 
	where $\rmU^{\prime}$ is column orthogonal referring to a \emph{rotation} of $\rmU$.
	Consequently, applying different $\rmS \in \gM_{G}$ all correspond to filtering $\vf$ with related filters 
	under different decomposition $\rmU^{\top}$~\footnote{Furthermore, $\rmS\vf = \rmU^{\prime} \hat{\rmLambda}^{\prime} (\rmU^{\prime\top} \vf) = \rmU^{\prime} (\hat{\vlambda}^{\prime} \odot (\rmU^{\prime\top} \vf)) = \rmU^{\prime} ((\rmU^{\prime\top} \vlambda^{\prime}) \odot (\rmU^{\prime\top} \vf)) = \vlambda^{\prime} *_{\rmS} \vf$, 
		which is consistent with the definition of spectral graph convolution.}.
	To build a generalized point of view, we generalize the spectral graph convolution in Eq.~\ref{equ:gc} as follows.
	\begin{definition}[Parameterized-$(\gD, \gF)$]
		\label{def:para_df}
		Given a graph $G$, and a signal $\vf$ assigned on $G$, 
		the Parameterized-$(\gD, \gF)$ of $\vf$ over $G$ is defined as
		\begin{equation}
			\label{equ:ggc}
			\begin{aligned}
				\vf^{\prime} = f_{\vtheta}(\gG) \vf
			\end{aligned}    
		\end{equation}
		where $\gG \subset \gM_{G}$ 
		and $f_{\vtheta}: \{ \sR^{n \times n} \} \mapsto \sR^{n \times n}$ is the mapping of $\gG$ with the parameter $\vtheta$. 
		$f_{\vtheta}$ should satisfy the conditions, 
		\begin{compactitem}
			\item{\emph{Closeness}:} $\forall \, \gG \subset \gM_{G} \,, \, f_{\vtheta}(\gG) \in \gM_{G}$, and
			\item{\emph{Permutation-equivariance:}} For any permutation matrix $\rmM \in \sR^{n \times n}$,
			$f_{\vtheta}(\{ \rmM^{\top} \rmS \rmM \, | \, \rmS \in \gG\}) = \rmM^{\top} f_{\vtheta}(\gG) \rmM$.
		\end{compactitem}
	\end{definition}
	Given $\gG$, $f_{\vtheta}(\gG)$ is parameterized by $\vtheta$, and the resulting space w.r.t. $\vtheta$ is $\{f_{\vtheta}(\gG)|\vtheta\} \subset \gM_{G}$.
	Any $\rmS \in \{f_{\vtheta}(\gG)|\vtheta\}$ corresponds to a specific $(\gD, \gF)_{\rmS}$.
	$\{g_{\vtheta}(\rmL)|\vtheta\in\mathbb R^k\}$ in Eq.\ref{equ:gc} based on the theory of spectral graph convolution 
	refers to a constrained Parameterized-$(\gD, \gF)$: 
	(i) $\gG = \{\rmL\}$ as required in Graph Fourier Transform, and 
	(ii) $f_{\vtheta} = g_{\vtheta}$ is constrained to be polynomials 
	such that it is equivariant to the decomposition $\rmU$, 
	i.e., $g_{\vtheta}(\rmU\rmLambda \rmU^{\top})=\rmU g_{\vtheta}(\rmLambda) \rmU^{\top}$.~\footnote{Or alternatively, 
		$\gG = \{\rmL^{k}|k\in[K]\}$, and $f_{\vtheta} = \sum_{k\in[K]} \vtheta_k \cdot$.}
	This makes all $\rmS \in \{g_{\vtheta}(\rmL)|\vtheta\}$ share the same $\gD$.
	In other words, applying learnable $\vtheta$ will only learn $\gF$ but fix $\gD$.
	
	By removing such constraints,
	different $\vtheta$ results in different $\gD$ and $\gF$.
	Therefore, applying learnable $\vtheta$ makes it capable of learning $(\gD, \gF)$ as a whole.
	We will systematically investigate the effectiveness of learnable $\gD$ 
	by first revisiting existing GNNs via the lens of Parameterized-$(\gD, \gF)$ in Sec.~\ref{sec:unifying}, 
	and then developing our models leveraging the learnable $\gD$ in Sec.~\ref{sec:implement}.
	In Sec.~\ref{sec:analysis}, we will conduct analysis on 
	the limitations of learning $\gF$ alone and the necessity of learning $(\gD, \gF)$ as a whole.
	
	\subsection{Unifying Existing GNNs with Parameterized-$(\gD, \gF)$}
	\label{sec:unifying}
	
\begin{figure}[t]
		\centering
		\includegraphics[width=\columnwidth]{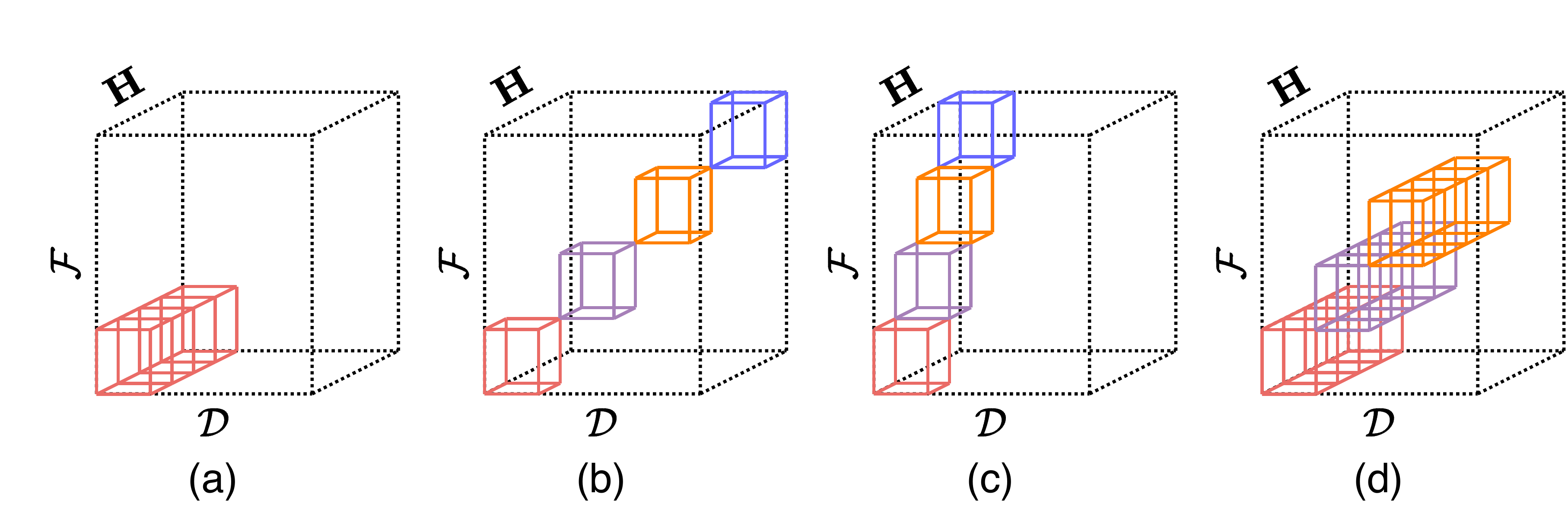}
		\vspace{-20pt}
		\caption{A taxonomy of existing GNN architectures via the lens of $(\gD, \gF)$ for multichannel signal $\rmH$. 
			Within each of them, the colored box corresponds to a channel processed with an individual $(\gD, \gF)$.
			We use the same color to denote the same $(\gD, \gF)$. 
			Each entry in the $\gD$-$\gF$ plane corresponds to a specific $(\gD, \gF)$ $\in\gM_{G}$. 
			As eigendecomposition is unique, these $(\gD, \gF)$ are different.}
		\label{fig:decomposition_filtering}
		\vspace{-10pt}
	\end{figure}

	By assigning constraints on $f_{\vtheta}$ and $\gG$ in Eq.~\ref{equ:ggc}, 
	we can achieve (fixed $\gD$, fixed $\gF$), (fixed $\gD$, learnable $\gF$) or (learnable $\gD$, learnable $\gF$) respectively, 
	which provides a unification of various GNNs.
	First, we summarize all possible architectures in multichannel signal scenario as in Fig.~\ref{fig:decomposition_filtering}.
	\begin{compactenum}[(a)]
		\item refers that all channels share the same $(\gD, \gF)$;
		\item assigns each channel with an independent $(\gD, \gF)$;
		\item applies independent $\gF$ under a shared $\gD$ in different channels;
		\item assigns each signal with multiple $(\gD, \gF)$.
	\end{compactenum}
	\begin{table*}[h]
		\small
		\centering
		\caption{A summary of differences of popular GNNs via the lens of $(\gD, \gF)$. 
			TF denotes Transformer and MGAT denotes multi-head GAT.}
		\label{tab:taxonomy}
		\vspace{5pt}
		\resizebox{0.95\textwidth}{!}{%
			\begin{tabular}{l|cccccccc}\toprule
				&GCN/SGC &BernNet/GPR &JacobiConv &Corr-free &PGSO/GAT &GIN-0 &TF/MGAT/ExpC &PNA\\ \midrule
				Category &Spectral &Spectral &Spectral &? &? &Spatial &Spatial &Spatial\\
				Learnable $\gF$ &No &Yes &Yes &Yes &Yes &No &Yes &No\\
				Learnable $\gD$ &No &No &No &Yes &Yes &No &Yes &No \\
				Architecture &(a) &(a) &(c) &(b)$^{\star}$ &(a) &(a) &(d) &(d)\\ \bottomrule
			\end{tabular}
		}
		\vspace{-10pt}
	\end{table*}
	Then, by integrating learnable $\gD$ or $\gF$ into each of these architectures, 
	we can classify all existing models as in Tab.~\ref{tab:taxonomy}.
	Note that (b) has not been well investigated yet, and to the best of our knowledge, 
	only Correlation-free~\citep{yang2022spectrum} implicitly results in this architecture 
	due to nonlinear computations in their code implementation, 
	but there is no reflection of this architecture in their paper.
	(a) acts as the most common form under which many studies correspond to 
	designing sophisticated polynomials to parameterize $\gF$, e.g., ChebyNet/CayleyNet/BernNet.
	JacobiConv differs from them in that it learns $\gF$ for each channel independently, therefore it corresponds to (c).
	Compared with (b), the channel-wise learnable $\gF$ in (c) is still under the shared $\gD$.
	(d) assigns each channel with multiple $(\gD, \gF)$.
	It is generalized from multi-head attention, 
	e.g., Multi-head GAT/Transformer, or multi-aggregator, e.g., PNA/ExpC, where each \emph{head} in multi-head attention, 
	or each \emph{aggregator} in multi-aggregator corresponds to an individual $(\gD, \gF)$.
	Their effectiveness can be interpreted by filtering task-relevant patterns from multiple $\gD$ for each input channel.
	Some spatial methods with learnable aggregation coefficients implicitly result in learnable $\gD$, e.g., GAT/ExpC/PNA, as the resulting $\rmS \in \gM_{G}$ with different aggregation coefficients generally do not share $\gD$.
	
	Eq.~\ref{equ:ggc} also inspires a new perspective regarding the difference between spectral and spatial methods~\footnote{There are no 
		formal definitions of spectral and spatial-based methods. 
		Generally, the ones introduced from spectral graph convolution 
		as in Eq.~\ref{equ:gc} are considered as spectral-based methods, 
		and others such as message-passing, Graph Transformers are considered as spatial ones.}: 
	if $\gD$ is fixed to be the eigenspace of Laplacian, they belong to spectral models, 
	and can be implemented with fixed/learnable or channel-shared/independent $\gF$ strategies.
	Otherwise, they belong to non-spectral or spatial models.
	
	\subsection{Developing an Effective Parameterized-$(\gD, \gF)$}
	\label{sec:implement}
	
	We have shown that some competitive graph-level prediction models implicitly include learnable $\gD$.
	But they are introduced as a side effect of various attention, neighborhood aggregation, and transformer designs.
	And their learning space (the set of all $\gD$ that can be learned from the given graph) 
	varies from each other due to different implementations.
	We believe that the learnable $(\gD, \gF)$ as a whole 
	can potentially contribute to the final performance improvements.
	To fully leverage learnable $\gD$ as well as $\gF$ for a graph $G$, 
	we develop Parameterized-$(\gD, \gF)$ with the objective of 
	learning $(\gD, \gF)$ from a larger subspace of $\gM_{G}$ for input graph signals.
	
	Note that $f_{\vtheta}(\gG)$ involves two components: $\gG$ and $f_{\vtheta}$.
	Building $\gG$ can be flexible.
	For example, in some attention-based methods, e.g., GAT and Graph Transformers, 
	$\gG$ is implemented as an attention weight matrix that takes advantage of both graph topology and node features.
	In PGSO, $\gG$ is a group of GSOs.
	Here, we build $\gG$ as $\gG=\{(\tilde \rmD^{\eps}\tilde \rmA\tilde \rmD^{\eps})^k|-0.5\leq\eps\leq 0, k\in[K]\}$ 
	in order that $\tilde \rmD^{\eps}\tilde \rmA\tilde \rmD^{\eps}$ with different $\eps$ do not share eigenspace.
	There can be more sophisticated designs of $\gG$, and we leave them for future work.
	Then, we implement $f_{\vtheta}$ as a multi-layer perceptron (MLP), 
	which satisfies the two conditions in Definition~\ref{def:para_df}.
	For the sake of brevity, we only present one layer perceptron here without loss of generality.
	Correspondingly, the convolution on a single-channel signal $\vf$ is
	\begin{equation}
		\label{equ:single_ggc}
		\begin{aligned}
			\vf^{\prime}=f_{\vtheta}(\gG)\vf=\sigma(\sum_{\rmS_i\in\gG} \vtheta_i \rmS_i)\vf,
		\end{aligned}
	\end{equation}
	where $\vtheta\in\mathbb R^{|\gG|}$ is the learnable coefficient and $\sigma$ is a nonlinear function in the MLP.
	Given $G$ and its associated $\gG$, the learning space of $(\gD, \gF)$ w.r.t. $\vtheta$ is $\{\sigma(\sum_{\rmS_i\in\gG} \vtheta_i \rmS_i)|\vtheta\}$.
	Applying more expressive $f_{\vtheta}$ allows learning $(\gD, \gF)$ from a larger subspace in $\gM_{G}$.
	In Sec.~\ref{sec:ablation_studies}, we conduct extensive experiments to 
	evaluate the effects of different $f_{\vtheta}$ and $\gG$.
	
	To extend Eq.~\ref{equ:single_ggc} to multichannel signal scenario, 
	we design Channel-shared Parameterized-$(\gD, \gF)$ (denoted by ``shd-\method'') and 
	Channel-independent Parameterized-$(\gD, \gF)$ (denoted by ``idp-\method'') 
	strategies with each of them divided into three steps, differing at Step 2.
	\begin{compactenum}[1)]
		\label{enum:update_rule}
		\item{\underline{Pre-transform}:} $\rmZ=\sigma(\rmH\rmW_1)$
		\item{\underline{Multichannel Parameterized-$(\gD, \gF)$}:}
		\begin{itemize}
			\item{\underline{shd-\method:}} $\rmZ^{\prime}=\sigma(\sum_{\rmS_i \in \mathcal{G}} \vtheta_i \rmS_i) \rmZ$
			\item{\underline{idp-\method:}} $\rmZ^{\prime}_{:j}=\sigma(\sum_{\rmS_i\in\mathcal{G}} \rmTheta_{ij} \rmS_i) \rmZ_{:j}$
		\end{itemize}
		\item{\underline{Post-transform}:} $\rmH^{\prime}=\sigma(\rmZ^{\prime} \rmW_2)$
	\end{compactenum}
	Correspondingly, shd-\method belongs to Fig.~\ref{fig:decomposition_filtering}(a), 
	and idp-\method belongs to Fig.~\ref{fig:decomposition_filtering}(b) which 
	has not been well investigated by existing studies.
	$\rmW_{1}, \rmW_{2} \in \mathbb R^{d\times d}$ are learnable transformation matrices.
	In shd-\method, the channel-shared $(\gD, \gF)$ is parameterized by the learnable $\vtheta \in \mathbb R^{|\gG|}$, 
	and in idp-\method, the channel-independent $(\gD, \gF)$ is parameterized by the learnable $\rmTheta \in \mathbb R^{|\gG|\times d}$.
	
	Compared with the latest best-performing methods, shd-\method and idp-\method benefit from their simple implementations and computational efficiency.
	The time and space complexity and the running time are provided in Appendix~\ref{app:complexity}.
	In our experiments, the performance improvements under these simple implementations validate the effectiveness of Parameterized-$(\gD, \gF)$.
	
	\textbf{Scalability.}
	As graph matrices are generally stored in a sparse manner, 
	which only stores non-zero entries to save memory space, the proportion of non-zero entries determines the amount of memory usage.
	Especially for large graphs, we need to leverage sparse matrix representations, which is a subset of graph matrix representations, to make it tractable under the limited memory resources.
	There are many ways to express larger scope substructure while preserving the matrix sparsity such as encoding local structures to graph matrix entries.
	They can all be viewed as a sparse subset of $\gM_{G}$, which is denoted as $\gM^{\textrm{sps}}_{G}$.

\begin{figure}[h]
		\centering
		\includegraphics[width=\columnwidth]{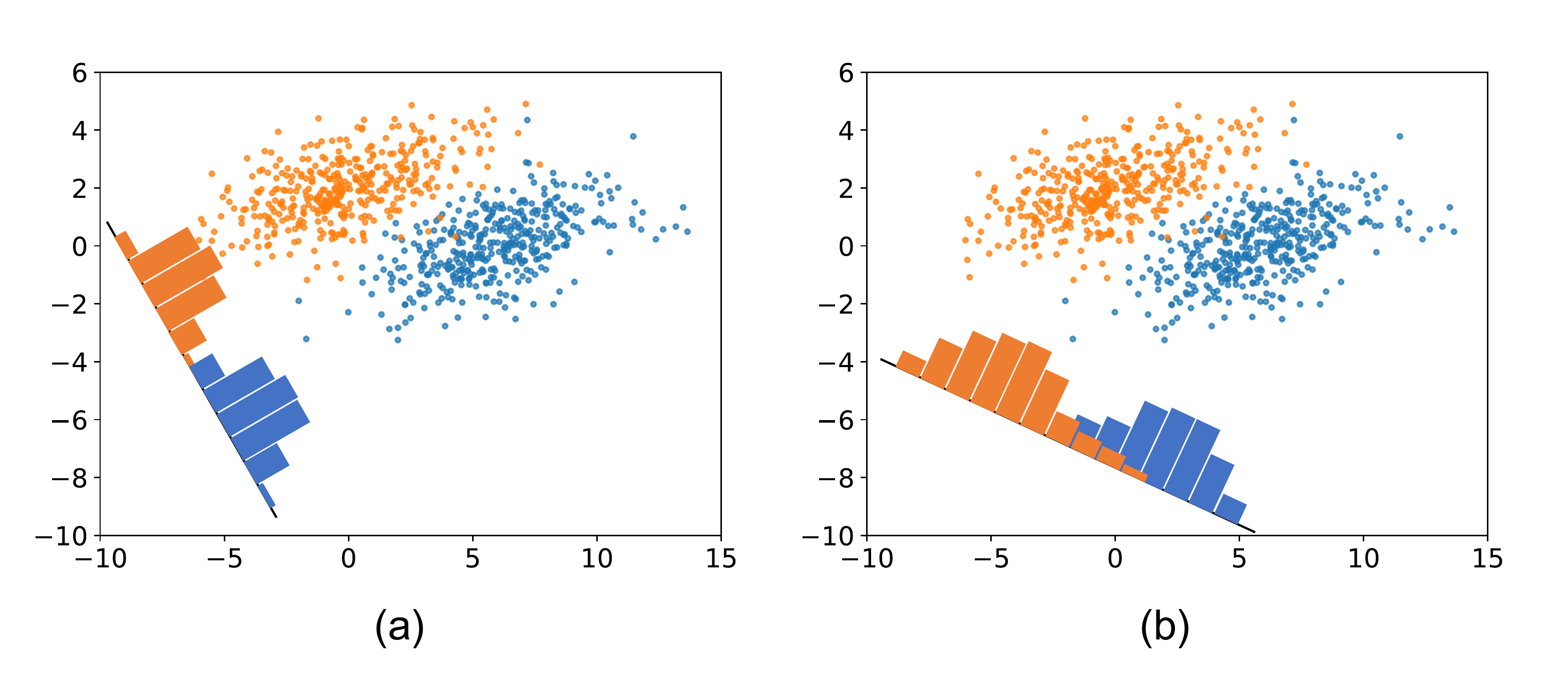}
		\vspace{-20pt}
		\caption{The effects of basis choices when considering within a bunch of signal channels.}
		\label{fig:rotation_demo}
		\vspace{-10pt}
	\end{figure}

	\textbf{Why is it necessary to learn $\gD$ from the input signals?}
	We provide an intuitive example in Fig.~\ref{fig:rotation_demo} here to illustrate the considerations.
	Suppose all signal channels form two clusters and the downstream task requires identifying these two clusters, then $\gD$ in Fig.~\ref{fig:rotation_demo}(a) serves as a more suitable one compared with that in Fig.~\ref{fig:rotation_demo}(b), as the projections on that basis better distinguish the two clusters.
	Therefore, the characteristics/distributions of input signals should be taken into consideration when choosing $\gD$, which serves as an empirical understanding of learnable $\gD$.
	This idea is well-adopted in dimension reduction techniques, e.g., PCA~\citep{abdi2010principal}, LDA~\citep{xanthopoulos2013linear}, etc.
	The theoretical motivations behind this are given in Sec.~\ref{sec:analysis}.

	\section{Motivations on Learning $(\gD, \gF)$ as a Whole}
	\label{sec:analysis}
	
	A multi-channel graph signal $\rmH \in \mathbb{R}^{n \times d}$ 
	involves $d$ channels.
	Along with the iterative graph convolution operations, 
	the smoothness of each channel dynamically change w.r.t. model depth, 
	and oversmoothed signals account for the performance drop of deep models.
	In this section, we study the smoothness of signals from two complementary perspectives, 
	i.e., the smoothness within a single channel and that over different channels, 
	showing how Parameterized-$(\gD, \gF)$ as a whole affects both of them.
	
	\subsection{Smoothness within a Single Channel}
	\label{sec:single_signal}
	
	For a weighted undirected graph $G$, its Laplacian is defined as $\rmL=\rmD-\rmW$, 
	where $\rmW$ is the weighted adjacency matrix with $\rmW_{ij}>0$ if vertices $i$ and $j$ are adjacent, and $\rmW_{ij}=0$ otherwise.
	$\rmD = \mathrm{diag}(\rmW \bm{1}_n)$ is the diagonal degree matrix.
	
	\textbf{Smoothness.}
	The smoothness of a graph signal $\vf\in\mathbb R^n$ w.r.t. $G$ is measured in terms of a quadratic form of the Laplacian~\citep{zhou2004regularization,shuman2013emerging}:
	\begin{equation}
		\label{equ:smoothness}
		\begin{aligned}
			\vf^{\top} \rmL \vf=\frac{1}{2} \sum_{i, j\in[n]} \rmW_{i j}(\vf(i)-\vf(j))^2,
		\end{aligned}
	\end{equation}
	where $\vf(i)$ and $\vf(j)$ are the signal values associated with these two vertices. 
	Intuitively, given that the weights are non-negative, 
	Eq.~\ref{equ:smoothness} shows that a graph signal $\vf$ is considered to be smooth 
	if strongly connected vertices (with a large weight on the edge between them) have similar values. 
	In particular, the smaller the quadratic form in Eq.~\ref{equ:smoothness}, 
	the smoother the signal on the graph~\citep{dong2016learning}.
	
	In addition to the traditional smoothness analysis, we provide a new insight based on $(\gD, \gF)$.
	Let $\hat {\vf}_{i}=\rmU_{i}^{\top} \vf\in\mathbb R$ be the $i$-th component after Graph Fourier Transform, then
	\begin{equation}
		\label{equ:weighted_inner_product}
		\begin{aligned}
			\vf^{\top} \rmL \vf&=\vf^{\top} \rmU \rmLambda \rmU^{\top} \vf\\
			&=(\rmU^{\top} \vf)^{\top} \rmLambda \rmU^{\top} \vf
			=\hat {\vf}^{\top} \rmLambda \hat {\vf}\\
			&=\sum_{i=1}^{n} \hat {\vf}_{i}^{2}\vlambda_{i}.
		\end{aligned}
	\end{equation}
	As $\vlambda_{i}\geq 0$, Eq.~\ref{equ:weighted_inner_product} shows that the smoothness of a graph signal $\vf$ w.r.t. Laplacian $\rmL$ in vertex domain is equivalent to the squares of its \emph{weighted norm} in frequency domain with the spectrum of $\rmL$ serving as weights.
	Also, Eq.~\ref{equ:weighted_inner_product} shows that the smoothness w.r.t. $\rmL$ is decided by both decomposition $\rmU$ and filtering $\rmLambda$.
	Next, we show what signal smoothness on the underlying graph implies in graph neural networks.
	
	The learnable polynomial filter designs are the most popular approaches in spectral GNN studies, e.g., ChebyNet~\citep{defferrard2016convolutional}, CayleyNet~\citep{ron2019cayleynets}, BernNet~\citep{he2021bernnet}, GPR~\citep{chien2021adaptive}, JacobiConv~\citep{JacobiConv}, Corr-free~\citep{yang2022spectrum}, etc.
	But applying various polynomials $g_{\vtheta}$ only introduces modifications on filtering, with decomposition unchanged as $g_{\vtheta}(\rmL)=\rmU g_{\vtheta}(\rmLambda)\rmU^{\top}$.
	This leads to the negative effect on signal smoothness as follows.
	\begin{proposition}
		\label{prop:smoothness}
		For a polynomial graph filter with the polynomial $g_{\vtheta}$ and the underlying (normalized) Laplacian $\rmL$ with the spectrum $\vlambda_{i}, i\in[n]$,\\
		(i) if $|g_{\vtheta}(\vlambda_{i})|<1$ for all $i\in[n]$, the resulting graph convolution \emph{smooths} any input signal w.r.t. $\rmL$;\\
		(ii) if $|g_{\vtheta}(\vlambda_{i})|>1$ for all $i\in[n]$, the resulting graph convolution \emph{amplifies} any input signal w.r.t. $\rmL$.
	\end{proposition}
	We prove Proposition~\ref{prop:smoothness} in Appendix~\ref{proof:prop:smoothness}.
	Proposition~\ref{prop:smoothness} shows that if we fix $\gD$ and consider $\gF$ alone, the smoothness of the resulting signals is sensitive to the range of $g_{\vtheta}(\vlambda_{i})$.
	A freely learned polynomial coefficient $\vtheta$ is more likely to result in the smoothness issue.
	Especially in deep models, the smoothness issue will accumulate.
	This may explain why learnable polynomial filter design is challenging and usually requires sophisticated polynomial bases.
	For example, JacobiConv uses Jacobi polynomials and configures the basis $P_k^{a, b}$ by carefully setting $a$ and $b$~\citep{JacobiConv}.
	
	Some methods alternatively use pre-defined filters to avoid learning a filter with the above smoothness issue~\citep{pmlr-v97-wu19e,klicpera_predict_2019,chenWHDL2020gcnii,klicpera2019diffusion,zhu2020simple}.
	However, for some models, the smoothness issue still exists.
	\begin{proposition}
		\label{prop:gcn_smoothness}
		The GCN layer always smooths any input signal w.r.t. $\tilde \rmL=\rmI-\tilde \rmD^{-\frac{1}{2}}\tilde \rmA\tilde \rmD^{-\frac{1}{2}}$.
	\end{proposition}
	We prove Proposition~\ref{prop:gcn_smoothness} in Appendix~\ref{proof:prop:gcn_smoothness}.
	Proposition~\ref{prop:gcn_smoothness} shows that the accumulation of signal smoothness may account for the performance degradation of deep GCN.
	Similarly, we can prove that SGC~\citep{pmlr-v97-wu19e} suffers from the same issue.
	Some other fixed filter methods apply residual connections and manually set spectrum shift, which implicitly help to preserve the range of $g_{\vtheta}(\vlambda_{i})$, making them tractable for stacking deep architectures~\citep{klicpera_predict_2019,chenWHDL2020gcnii,xu2018representation}.
	
	\textbf{Comparisons with existing smoothness analysis.}
	First, based on the definition of smoothness in Eq.~\ref{equ:smoothness}, the smoothness analysis of signals in Proposition~\ref{prop:smoothness} and Proposition~\ref{prop:gcn_smoothness} is always considered with the underlying graphs, while the convergence analysis in most existing oversmoothing studies is not considered under the graph topology.
	Also, they only deal with the theoretical infinite depth case and study the final convergence result.
	Second, Proposition~\ref{prop:smoothness} shows that in addition to smoothness issue, the amplification issue will also occur complementarilly with inappropriate filter design.
	The amplification issue is less discussed, in contrast, the numerical instabilities as a reflection of the amplification issue are more well-known.
	And most methods choose $\tilde \rmD^{-\frac{1}{2}}\tilde \rmA\tilde \rmD^{-\frac{1}{2}}$ with the bounded spectrum $[-1, 1]$ to avoid numerical instability~\citep{kipf2017semi}.
	Some other studies find the signal amplification issue from their empirical evaluations and overcome it by proposing various normalization techniques~\citep{zhou2021understanding,guo2022orthogonal}.
	Our analysis shows that both smoothness and amplification issues can somehow appear complementarily in (fixed $\gD$, learnable $\gF$) designs for deeper models, which reveals the connections between existing oversmoothing and numerical instability/amplification studies.
	
	
	\subsection{Smoothness over Different Channels}
	\label{sec:bunch_signal}
	Apart from the smoothness within each channel, the smoothness over different channels in hidden features also indicates information loss and affects the performance.
	For example, the cosine similarity between signal pairs is usually used as a metric of smoothness over different channels.
	The larger the cosine similarity is, the smoother they are to each other.
	In the extreme case, all signal pairs have a cosine similarity equal to one, 
	which means the worst information loss case where each signal is linearly dependent on each other.
	\cite{yang2022spectrum} shows that in a polynomial graph filter with the resulting matrix representation $\rmS=\rmU g_{\vtheta}(\rmLambda) \rmU^{\top}$,
	\begin{equation}
		\label{equ:cos_signals}
		\begin{aligned}
			\cos \left(\left\langle\rmS\vf, \rmU_i\right\rangle\right)=\frac{(\rmU_i^{\top} \vf) g_{\vtheta}(\vlambda_{i})}{\sqrt{\sum_{j=1}^n (\rmU_j^{\top} \vf)^2 g_{\vtheta}(\vlambda_{j})^2}}.
		\end{aligned}
	\end{equation}
	As the model goes deeper, the spectrum diversity will accumulate exponentially, i.e. $\rmS^k=\rmU g_{\vtheta}(\rmLambda)^k \rmU^{\top}$.
	As a result, all signals will tend to correlate to the leading eigenvector $\rmU_0$.
	Inspired by this, some work explores strategies to restrict the expanding of spectrum diversity in deep models to alleviate this issue, but in return restricting the expressiveness of polynomial filters~\citep{yang2022spectrum}.
	
	However, note that
	$\rmU_i^{\top}\vf$ and $g_{\vtheta}(\vlambda_{i})$ have equivalent contributions to the smoothness over different channels as reflected in Eq.~\ref{equ:cos_signals}.
	We can explore a better matrix representation, which improves $(\gD, \gF)$ as a whole, to avoid restrictions imposed on the filters as introduced in other works. 
	
	The smoothness of a single channel and that over multichannels serve as two orthogonal perspectives, having been investigated by many existing works, e.g., the former studied by~\cite{li2018deeper,oono2020graph,Rong2020DropEdge:,huang2020tackling}, and the latter studied by~\cite{zhao2020pairnorm,liu2020towards,chien2021adaptive,yang2022spectrum,jin2022feature}.
	In comparison to existing work which only consider $\gF$ and related spectrum, we further involve $\gD$, indicating that by treating $(\gD, \gF)$ as a whole, we can handle the above smoothness issues simultaneously. 
	
	\section{Experiments}
	We evaluate our model \method induced from Parameterized-$(\gD, \gF)$ on the graph-level prediction tasks.
	Detailed information of the datasets is given in Appendix~\ref{appe:dataset_statistics}.
	We first conduct extensive ablation studies to validate its effectiveness, and then compare its performance with baselines.
	
	\subsection{Ablation Studies}
	\label{sec:ablation_studies}
	We evaluate the effectiveness of \method as the instantiation of Parameterized-$(\gD, \gF)$ on multichannel signal scenario on ZINC following its default dataset settings.
	We use ``shd" and ``idp" to represent the channel-shared and channel-independent architectures, respectively.
	Both architectures learn the corresponding coefficients $\vtheta \in \mathbb R^{|\gG|}$ and $\rmTheta \in \mathbb R^{|\gG|\times d}$ from scratch.
	
	Parameterized-$(\gD, \gF)$ involves two components: the construction of $\gG$ and the implementation of $f_{\vtheta}$.
	For each of them, we design the following variants:
	\begin{compactitem}
		\item For the matrix representation $\gG$,
		\begin{compactitem}
			\item Lap: $\gG=\{(\tilde \rmD^{-\frac{1}{2}}\tilde \rmA\tilde \rmD^{-\frac{1}{2}})^k|k\in[K]\}$,
			\item $(\eps, k)$: $\gG=\{(\tilde \rmD^{\eps}\tilde \rmA\tilde \rmD^{\eps})^k|-0.5\leq\eps\leq 0, k\in[K]\}$;
		\end{compactitem}
		\item For the mapping function $f_{\vtheta}$,
		\begin{compactitem}
			\item Lin: $f_{\vtheta}$ is a learnable affine transformation,
			\item 1L: $f_{\vtheta}$ is a 1-layer perceptron,
			\item 2L: $f_{\vtheta}$ is a 2-layer perceptron.
		\end{compactitem}
	\end{compactitem}
	$(\textrm{Lap}, \textrm{Lin})$ corresponds to general spectral GNNs that apply fixed normalized Laplacian decomposition and learnable filtering.
	$((\eps, k), \textrm{Lin})$, $((\eps, k), \textrm{1L})$ and $((\eps, k), \textrm{2L})$ belong to learnable $(\gD, \gF)$.
	shd-$((\eps, k), \textrm{Lin})$ is similar to PGSO with the involved $\gG$ slightly different.
	$((\eps, k), \textrm{2L})^{\textrm{sps}}$ is used to test the effects of learning within sparse representation subspace $\gM^{\textrm{sps}}_{G}\subset\gM_{G}$, 
	which is implemented by masking all neighbors beyond 2-hop in $((\eps, k), \textrm{2L})$.
	Detailed model configurations of each test case are provided in Appendix~\ref{appe:ablation_setting}.
	Tab.~\ref{tab:ablation} presents the ablation study results, and more detailed statistics are in Appendix~\ref{appe:more_results}.
	\begin{table}[t]
		\centering
		\caption{Ablation studies on ZINC.}
		\label{tab:ablation}
		\vspace{5pt}
		\resizebox{0.9\columnwidth}{!}{%
			\begin{tabular}{ll|cc}
				\toprule
				\multicolumn{2}{c}{Ablation} &valid MAE &test MAE\\
				\midrule
				\multirow{5}{*}{shd+}&$(\textrm{Lap}, \textrm{Lin})$ &0.227$\pm$0.0445 &0.219$\pm$0.0520\\
				&$((\eps, k), \textrm{Lin})$ &0.184$\pm$0.0276 &0.167$\pm$0.0234\\
				&$((\eps, k), \textrm{2L})^{\textrm{sps}}$ &0.174$\pm$0.0125 &0.150$\pm$0.0141\\
				&$((\eps, k), \textrm{1L})$ &0.172$\pm$0.0087 &0.160$\pm$0.0119\\
				&$((\eps, k), \textrm{2L})$ &0.121$\pm$0.0137 &0.112$\pm$0.0138\\
				\midrule
				\multirow{5}{*}{idp+}&$(\textrm{Lap}, \textrm{Lin})$ &0.188$\pm$0.0048 &0.172$\pm$0.0041\\
				&$((\eps, k), \textrm{Lin})$ &0.168$\pm$0.0071 &0.150$\pm$0.0038\\
				&$((\eps, k), \textrm{2L})^{\textrm{sps}}$ &0.127$\pm$0.0028 &0.111$\pm$0.0024\\
				&$((\eps, k), \textrm{1L})$ &0.104$\pm$0.0028 &0.088$\pm$0.0031\\
				&$((\eps, k), \textrm{2L})$ &\textbf{0.085}$\pm$\textbf{0.0038} &\textbf{0.066}$\pm$\textbf{0.0020}\\
				\bottomrule
			\end{tabular}
		}
		\vspace{-10pt}
	\end{table}
	
	\underline{\emph{Effectiveness of learnable $(\gD, \gF)$}:}
	$(\textrm{Lap}, \textrm{Lin})$ is analogous to spectral graph convolutions where all matrix representations that can be learned share the same eigenspace, which results in a fixed $\gD$.
	Also, Lap refers to $(-0.5, k) \subset (\eps, k)$, and therefore, for a given $G$ and Lin, 
	the learning space w.r.t. $\rmTheta$ has $\{((-0.5, k), \textrm{Lin})|\rmTheta\} \subset \{((\eps, k), \textrm{Lin})|\rmTheta\}$.
	The results show that $((\eps, k), \textrm{Lin})$ with learnable $(\gD, \gF)$ 
	outperforms $(\textrm{Lap}, \textrm{Lin})$ with learnable $\gF$  on both architectures.
	
	\underline{\emph{Effects of the expressiveness of $f_{\vtheta}$}:}
	Note that the expressiveness comparisons of $f_{\vtheta}$ in $((\eps, k), \textrm{Lin})$, $((\eps, k), \textrm{1L})$ 
	and $((\eps, k), \textrm{2L})$ is $\textrm{Lin} \prec \textrm{1L} \prec \textrm{2L}$.
	Thus, for a given graph $G$ and $(\eps, k)$, the learning space comparison w.r.t. $\rmTheta$ is
	\begin{equation}
		\nonumber
		\begin{aligned}
			\{((\eps, k), \textrm{Lin})|\rmTheta\} &\subset \{((\eps, k), \textrm{1L})|\rmTheta\} \\
			&\subset \{((\eps, k), \textrm{2L})|\rmTheta\} \subset \gM_{G}.
		\end{aligned}
	\end{equation}
	This is exactly reflected on their performance comparisons where the more expressive $f_{\vtheta}$ is, 
	the better performance the resulting model achieves.
	This validates our analysis that learning from a larger subspace within $\gM_{G}$ helps to 
	find more effective graph matrix representation, i.e., $(\gD, \gF)$ for input signals.
	Also, 2L is sufficient to be used to parameterize any desired mapping according to the universal approximation theorem~\citep{hornik1989multilayer}, 
	and we can see that $((\eps, k), \textrm{2L})$ does benefit from this guarantee, 
	making it outperform the other two by a large margin.
	
	\underline{\emph{Effects of sparse representations}:}
	Compared with $((\eps, k), \textrm{2L})$, the learning space of $((\eps, k), \textrm{2L})^{\textrm{sps}}$ 
	is limited within $\gM^{\textrm{sps}}_{G}$, 
	and the results show that $((\eps, k), \textrm{2L})^{\textrm{sps}}$ does not outperform $((\eps, k), \textrm{2L})$.
	Sparse representations only correspond to a restricted subspace $\gM^{\textrm{sps}}_{G}\subset \gM_{G}$. It acts as a trade-off between scalability and prediction performance. Users are suggested to use dense representations if it is tractable on the given graph scales.

	\underline{\emph{Effects of architecture designs}:}
	Channel-independent learnable $(\gD, \gF)$ outperforms the channel-shared one in all cases.
	Meanwhile, channel-independent one 
	is much more stable, as reflected by smaller variations (STD) in different runs
	in Tab.~\ref{tab:ablation}, as well as the training curves in Appendix~\ref{appe:more_results}.
	These results empirically show that assigning each channel with an individual $(\gD, \gF)$ is more appropriate.

	\subsection{Performance Comparison}
	\label{sec:results}
	
	All baseline results of ZINC, ogbg-molpcba and pgbg-ppa in Tab.~\ref{tab:zinc_pcba_results} 
	are quoted from their leaderboards\footnote{\small{\url{https://paperswithcode.com/sota/graph-regression-on-zinc-500k}} and \small{\url{https://ogb.stanford.edu/docs/leader\_graphprop/}}} or the original papers.
	And all baseline results of TUDatasets in Tab.~\ref{tab:tu_results} are quoted from their original papers.
	All baseline models in this section are summarized in Appendix~\ref{appe:baseline}.
	Ogbg-ppa and RDT-B are large scale graphs with more than 200 vertices. 
	We apply sparse matrix representations subspace $\gM^{\textrm{sps}}_{G}$ on them with the results marked with $*$.
	All detailed hyperparameter configurations can be found in Appendix~\ref{appe:setup}.
	\method appearing in the baseline comparisons refers to the implementation idp-$((\eps, k), \textrm{2L})$.
	\begin{table}[t]
		\centering
		\caption{Results on ZINC, ogbg-molpcba and ogbg-ppa with the number of parameters used, 
			where the best results are in bold, and second-best are underlined.}
		\label{tab:zinc_pcba_results}
		\vspace{5pt}
		\resizebox{\columnwidth}{!}{%
			\begin{tabular}{l|ccc}
				\toprule
				\multirow{2}{*}{Method} &ZINC &ogbg-molpcba &ogbg-ppa\\
				&$\textrm{MAE}\downarrow_{\#\textrm{para}}$  &$\textrm{AP(\%)}\uparrow_{\#\textrm{para}}$  &$\textrm{ACC(\%)}\uparrow_{\#\textrm{para}}$\\
				\midrule
				GCN &0.367$\pm$0.011 $_{505\mathrm k}$ &24.24$\pm$0.34 $_{2.0\mathrm m}$ &68.39$\pm$0.84 $_{0.5\mathrm m}$\\
				GIN &0.526$\pm$0.051 $_{510\mathrm k}$ &27.03$\pm$0.23 $_{3.4\mathrm m}$ &68.92$\pm$1.00 $_{1.9\mathrm m}$\\
				GAT &0.384$\pm$0.007 $_{531\mathrm k}$ &- &-\\
				GraphS &0.398$\pm$0.002 $_{505\mathrm k}$ &- &-\\
				GatedG &0.214$\pm$0.006 $_{505\mathrm k}$ &- &-\\
				MPNN &0.145$\pm$0.007 $_{481\mathrm k}$ &- &-\\
				DeeperG &- &28.42$\pm$0.43 $_{5.6\mathrm m}$ &77.12$\pm$0.71 $_{2.3\mathrm m}$\\
				PNA &0.142$\pm$0.010 $_{387\mathrm k}$ &28.38$\pm$0.35 $_{6.6\mathrm m}$ &-\\
				DGN &0.168$\pm$0.003 $_{\mathrm{NA}}$ &28.85$\pm$0.30 $_{6.7\mathrm m}$ &-\\
				GSN &0.101$\pm$0.010 $_{523\mathrm k}$ &- &-\\
				GINE-{\scriptsize AP} &- &\underline{29.79$\pm$0.30} $_{6.2\mathrm m}$ &-\\
				PHC-GN &- &29.47$\pm$0.26 $_{1.7\mathrm m}$ &-\\
				ExpC &- &23.42$\pm$0.29 $_{\mathrm{NA}}$ &79.76$\pm$0.72 $_{1.4\mathrm m}$\\
				\midrule
				GT &0.226$\pm$0.014 $_{\mathrm{NA}}$ &- &-\\
				SAN &0.139$\pm$0.006 $_{509\mathrm k}$ &27.65$\pm$0.42 $_{\mathrm{NA}}$ &-\\
				Graphor &0.122$\pm$0.006 $_{489\mathrm k}$ &- &-\\
				KS-SAT &0.094$\pm$0.008 $_{\mathrm{NA}}$ &- &75.22$\pm$0.56 $_{\mathrm{NA}}$\\
				GPS &\underline{0.070$\pm$0.004} $_{424\mathrm{k}}$ &29.07$\pm$0.28 $_{9.7\mathrm m}$ &\textbf{80.15}$\pm$\textbf{0.33} $_{3.4\mathrm m}$\\
				GM-Mix &0.075$\pm$0.001 $_{\mathrm{NA}}$ &- &-\\
				\midrule
				\method\small{(our)} &\textbf{0.066}$\pm$\textbf{0.002} $_{500\mathrm k}$ &\textbf{30.31}$\pm$\textbf{0.26} $_{3.8\mathrm m}$ &\underline{80.10$\pm$0.52}$^{*}$ $_{2.0\mathrm m}$\\
				\bottomrule
			\end{tabular}
		}
		\vspace{-15pt}
	\end{table}
	\begin{table*}[h]
		\centering
		\caption{Results on TUDataset (Higher is better).}
		\label{tab:tu_results}
		\vspace{5pt}
		\resizebox{0.95\textwidth}{!}{%
			\begin{tabular}{l|cccccccc}
				\toprule
				Method    & MUTAG &NCI1 &NCI109 &ENZYMES &PTC\_MR &PROTEINS &IMDB-B &RDT-B\\
				\midrule
				GK &81.52$\pm$2.11 &62.49$\pm$0.27 &62.35$\pm$0.3 &32.70$\pm$1.20 &55.65$\pm$0.5 &71.39$\pm$0.3 &- &77.34$\pm$0.18\\
				RW &79.11$\pm$2.1 &- &- &24.16$\pm$1.64 &55.91$\pm$0.3 &59.57$\pm$0.1 &- &-\\
				PK &76.0$\pm$2.7 &82.54$\pm$0.5 &- &- & 59.5$\pm$2.4 &73.68$\pm$0.7 &- &-\\
				FGSD &\textbf{92.12} &79.80 &78.84 &- &62.8 &73.42 &73.62 &-\\
				AWE &87.87$\pm$9.76 &- &- &35.77$\pm$5.93 &- &- &74.45$\pm$5.80 &87.89$\pm$2.53\\
				\midrule
				DGCNN &85.83$\pm$1.66 &74.44$\pm$0.47 &- &51.0$\pm$7.29 &58.59$\pm$2.5 &75.54$\pm$0.9 &70.03$\pm$0.90 &-\\
				PSCN &88.95$\pm$4.4 &74.44$\pm$0.5 &- &- &62.29$\pm$5.7 &75$\pm$2.5 &71$\pm$2.3 &86.30$\pm$1.58\\
				DCNN &- &56.61$\pm$1.04 &- &- &- &61.29$\pm$1.6 &49.06$\pm$1.4 &-\\
				ECC &76.11 &76.82 &75.03 &45.67 &- &- &- &-\\
				DGK &87.44$\pm$2.72 &80.31$\pm$0.46 &80.32$\pm$0.3 &53.43$\pm$0.91 &60.08$\pm$2.6 &75.68$\pm$0.5 &66.96$\pm$0.6 &78.04$\pm$0.39\\
				GraphSAGE &85.1$\pm$7.6 &76.0$\pm$1.8 &- &58.2$\pm$6.0 &- &- &72.3$\pm$5.3 &-\\
				CapsGNN &88.67$\pm$6.88 &78.35$\pm$1.55 &- &54.67$\pm$5.67 &- &76.2$\pm$3.6 &73.1$\pm$4.8 &-\\
				DiffPool &- &76.9$\pm$1.9 &- &\underline{62.53} &- &\textbf{78.1} &- &-\\
				GIN &89.4$\pm$5.6 &82.7$\pm$1.7 &- &- &64.6$\pm$7.0 &76.2$\pm$2.8 &\underline{75.1$\pm$5.1} &\underline{92.4$\pm$2.5}\\
				$k$-GNN &86.1 &76.2 &- &- &60.9 &75.5 &74.2 &-\\
				IGN &83.89$\pm$12.95 &74.33$\pm$2.71 &72.82$\pm$1.45 &- &58.53$\pm$6.86 &76.58$\pm$5.49 &72.0$\pm$5.54 &-\\
				PPGNN &\underline{90.55$\pm$8.7} &\underline{83.19$\pm$1.11} &82.23$\pm$1.42 &- &66.17$\pm$6.54 &\underline{77.20$\pm$4.73} &73.0$\pm$5.77 &-\\
				GCN$^2$ &89.39$\pm$1.60 &82.74$\pm$1.35 &\underline{83.00$\pm$1.89} &- &\underline{66.84$\pm$1.79} &71.71$\pm$1.04 &74.80$\pm$2.01 &-\\
				\midrule
				\method (ours) &89.91$\pm$4.35 &\textbf{85.47}$\pm$\textbf{1.38} &\textbf{83.62}$\pm$\textbf{1.38} &\textbf{73.50}$\pm$\textbf{6.39} &\textbf{68.36}$\pm$\textbf{8.38} &76.28$\pm$5.1 &\textbf{75.60}$\pm$\textbf{2.69} &\textbf{93.40}$\pm$\textbf{1.30}$^{*}$\\
				\bottomrule
			\end{tabular}
		}
		\vspace{-10pt}
	\end{table*}
	
	\textbf{ZINC.}
	We use the default dataset splits for ZINC, and following baselines settings on leaderboard, 
	set the number of parameters around 500K.
	\method achieves the superior performance on ZINC compared with all existing models,
	including both the lowest MAE and STD in multiple runs.
	
	\textbf{OGB.}
	We use the default dataset splits provided in OGB.
	The results in Tab.~\ref{tab:tu_results} involve molecular benchmark ogbg-molpcba with small and sparse connected graphs, 
	and protein-protein interaction benchmark ogbg-ppa with large and densely connected graphs.
	\method achieves the best AP with relatively fewer parameters on ogbg-molpcba.
	On ogbg-ppa, \method explores within $\gM^{\textrm{sps}}_{G}$ to balance computational efficiency, 
	and is still comparable to SOTA.
	
	\textbf{TUDataset.}
	We test our model on 8 TUDataset datasets involving both bioinformatics datasets 
	(MUTAG, NCI1, NCI109, ENZYMES, PTC\_MR and PROTEINS), and social network datasets (IMDB-B and RDT-B).
	To ensure a fair comparison with baselines, we follow the standard 10-fold cross-validation and 
	dataset splits in \cite{zhang2018end}, and then report our results according to the protocol described in \cite{xu2018how,ying2018hierarchical}.
	The results are presented in Tab.\ref{tab:tu_results}.
	\method achieves the highest classification accuracies on 6 out of 8 datasets, 
	among which \method outperforms existing models by a large margin on NCI1 and ENZYMES respectively.
	
	Also, \method benefits from its simple architecture and is more computational efficient than other SOTA models.
	In Appendix~\ref{app:complexity}, we show that \method shares a similar time complexity with GIN and GCN.
	Also, in our tests, the practical training and evaluation time on each epoch is similar to GIN, 
	which can be more than $2 \times$ faster than some latest SOTA models that improve their performance by 
	leveraging high expressive power or transformer architectures. 

	\section{Related Work}
	\label{sec:related_work}
	
	GWNN~\citep{xu2018graph} replaces the Fourier basis with wavelet basis, 
	which refers to a different decomposition on input signals.
	But similar to Fourier basis, it still uses fixed decomposition and cannot adopt relevant bases for input signals.
	
	JacobiConv~\citep{JacobiConv} and Corr-free~\citep{yang2022spectrum} learn individual filtering for each channel, 
	which is similar to our channel-independent architecture, but share the decomposition over these channels
	since they are induced by the spectral graph convolution theories.
	
	Parameterized Graph Shift Operator (PGSO)~\citep{dasoulas2021learning} is motivated by 
	spanning the space of commonly used GSOs as a replacement of the (normalized) Laplacian/Adjacency.
	As different GSOs generally do not share the eigenspace, PGSO can learn both $\gD$ and $\gF$.
	However, the linear combination of several GSOs can only leverage limited subspace within $\gM_{G}$.
	From the perspective of Parameterized-$(\gD, \gF)$, PGSO is somehow a restricted implementation of our shd-\method, 
	where $k$ is fixed to 1 and $f_{\vtheta}$ is a learnable linear transformation.
	The limited learning space may not be sufficient to obtain a suitable $(\gD, \gF)$ for input signals.
	As PGSO is analogous to one test case in our ablation studies, we can see that our model outperforms it,
	as shown in Sec.~\ref{sec:ablation_studies}.
	
	Multi-graph convolutions~\citep{geng2019spatiotemporal,khan2019multi} learn from multiple graphs, 
	with each one having its own semantic meaning.
	These methods are more application-oriented, e.g., traffic forecasting, and usually need domain expertise to define multigraphs, while our method has no such restriction.
	
	
	\section{Conclusion}
	
	In this work, we propose Parameterized-$(\gD, \gF)$, 
	which aims to learn a (decomposition, filtering) as a whole, i.e., a graph matrix representation, for input graph signals.
	It well unifies existing GNN models and the inspired new model achieves superior performance 
	while preserving computational efficiency.


	\section*{Acknowledgements}
	
	This project is supported by the National Natural Science Foundation of China under Grant 62276044, the National Research Foundation Singapore and DSO National Laboratories under the AlSingapore Programme (AISG Award No: AISG2-RP-2020-018), NUS-NCS Joint Laboratory (A-0008542-00-00), and CAAI-Huawei MindSpore Open Fund.

	
	\bibliographystyle{icml2023}
	\bibliography{reference}

\begin{thebibliography}{82}
\providecommand{\natexlab}[1]{#1}
\providecommand{\url}[1]{\texttt{#1}}
\expandafter\ifx\csname urlstyle\endcsname\relax
  \providecommand{\doi}[1]{doi: #1}\else
  \providecommand{\doi}{doi: \begingroup \urlstyle{rm}\Url}\fi

\bibitem[Abdi \& Williams(2010)Abdi and Williams]{abdi2010principal}
Abdi, H. and Williams, L.~J.
\newblock Principal component analysis.
\newblock \emph{Wiley interdisciplinary reviews: computational statistics},
  2\penalty0 (4):\penalty0 433--459, 2010.

\bibitem[Atwood \& Towsley(2016)Atwood and Towsley]{atwood2016diffusion}
Atwood, J. and Towsley, D.
\newblock Diffusion-convolutional neural networks.
\newblock In \emph{Advances in Neural Information Processing Systems}, pp.\
  1993--2001, 2016.

\bibitem[Bastos et~al.(2022)Bastos, Nadgeri, Singh, Kanezashi, Suzumura, and
  Mulang']{bastos2022how}
Bastos, A., Nadgeri, A., Singh, K., Kanezashi, H., Suzumura, T., and Mulang',
  I.~O.
\newblock How expressive are transformers in spectral domain for graphs?
\newblock \emph{Transactions on Machine Learning Research}, 2022.
\newblock URL \url{https://openreview.net/forum?id=aRsLetumx1}.

\bibitem[Beani et~al.(2021)Beani, Passaro, L{\'e}tourneau, Hamilton, Corso, and
  Li{\`o}]{beani2021directional}
Beani, D., Passaro, S., L{\'e}tourneau, V., Hamilton, W., Corso, G., and
  Li{\`o}, P.
\newblock Directional graph networks.
\newblock In \emph{International Conference on Machine Learning}, pp.\
  748--758. PMLR, 2021.

\bibitem[Bouritsas et~al.(2020)Bouritsas, Frasca, Zafeiriou, and
  Bronstein]{bouritsas2020improving}
Bouritsas, G., Frasca, F., Zafeiriou, S., and Bronstein, M.~M.
\newblock Improving graph neural network expressivity via subgraph isomorphism
  counting.
\newblock \emph{arXiv preprint arXiv:2006.09252}, 2020.

\bibitem[Bresson \& Laurent(2017)Bresson and Laurent]{bresson2017residual}
Bresson, X. and Laurent, T.
\newblock Residual gated graph convnets.
\newblock \emph{arXiv preprint arXiv:1711.07553}, 2017.

\bibitem[Brossard et~al.(2020)Brossard, Frigo, and Dehaene]{brossard2020graph}
Brossard, R., Frigo, O., and Dehaene, D.
\newblock Graph convolutions that can finally model local structure.
\newblock \emph{arXiv preprint arXiv:2011.15069}, 2020.

\bibitem[Butler \& Chung(2017)Butler and Chung]{butler2017spectral}
Butler, S. and Chung, F.
\newblock Spectral graph theory. handbook of linear algebra (2nd edition, l.
  hogben, ed.).
\newblock \emph{Discrete Mathematics and its Applications. CRC Press, Boca
  Raton}, pp.\  47/1—47/14, 2017.

\bibitem[Chen et~al.(2022)Chen, O’Bray, and Borgwardt]{chen2022structure}
Chen, D., O’Bray, L., and Borgwardt, K.
\newblock Structure-aware transformer for graph representation learning.
\newblock In \emph{International Conference on Machine Learning}, pp.\
  3469--3489. PMLR, 2022.

\bibitem[Chiang et~al.(2019)Chiang, Liu, Si, Li, Bengio, and
  Hsieh]{chiang2019cluster}
Chiang, W.-L., Liu, X., Si, S., Li, Y., Bengio, S., and Hsieh, C.-J.
\newblock Cluster-gcn: An efficient algorithm for training deep and large graph
  convolutional networks.
\newblock In \emph{Proceedings of the 25th ACM SIGKDD international conference
  on knowledge discovery \& data mining}, pp.\  257--266, 2019.

\bibitem[Chien et~al.(2021)Chien, Peng, Li, and Milenkovic]{chien2021adaptive}
Chien, E., Peng, J., Li, P., and Milenkovic, O.
\newblock Adaptive universal generalized pagerank graph neural network.
\newblock In \emph{International Conference on Learning Representations}, 2021.
\newblock URL \url{https://openreview.net/forum?id=n6jl7fLxrP}.

\bibitem[Chung(1997)]{chung1997spectral}
Chung, F.~R.
\newblock \emph{Spectral graph theory}, volume~92.
\newblock American Mathematical Soc., 1997.

\bibitem[Corso et~al.(2020)Corso, Cavalleri, Beaini, Li\`{o}, and
  Veli\v{c}kovi\'{c}]{corso2020pna}
Corso, G., Cavalleri, L., Beaini, D., Li\`{o}, P., and Veli\v{c}kovi\'{c}, P.
\newblock Principal neighbourhood aggregation for graph nets.
\newblock In \emph{Advances in Neural Information Processing Systems}, 2020.

\bibitem[Dasoulas et~al.(2021)Dasoulas, Lutzeyer, and
  Vazirgiannis]{dasoulas2021learning}
Dasoulas, G., Lutzeyer, J.~F., and Vazirgiannis, M.
\newblock Learning parametrised graph shift operators.
\newblock In \emph{International Conference on Learning Representations}, 2021.
\newblock URL \url{https://openreview.net/forum?id=0OlrLvrsHwQ}.

\bibitem[de~Haan et~al.(2020)de~Haan, Cohen, and Welling]{de2020natural}
de~Haan, P., Cohen, T.~S., and Welling, M.
\newblock Natural graph networks.
\newblock \emph{Advances in Neural Information Processing Systems},
  33:\penalty0 3636--3646, 2020.

\bibitem[Defferrard et~al.(2016)Defferrard, Bresson, and
  Vandergheynst]{defferrard2016convolutional}
Defferrard, M., Bresson, X., and Vandergheynst, P.
\newblock Convolutional neural networks on graphs with fast localized spectral
  filtering.
\newblock \emph{Advances in neural information processing systems},
  29:\penalty0 3844--3852, 2016.

\bibitem[Deri \& Moura(2017)Deri and Moura]{deri2017spectral}
Deri, J.~A. and Moura, J.~M.
\newblock Spectral projector-based graph fourier transforms.
\newblock \emph{IEEE Journal of Selected Topics in Signal Processing},
  11\penalty0 (6):\penalty0 785--795, 2017.

\bibitem[Dong et~al.(2016)Dong, Thanou, Frossard, and
  Vandergheynst]{dong2016learning}
Dong, X., Thanou, D., Frossard, P., and Vandergheynst, P.
\newblock Learning laplacian matrix in smooth graph signal representations.
\newblock \emph{IEEE Transactions on Signal Processing}, 64\penalty0
  (23):\penalty0 6160--6173, 2016.

\bibitem[Duvenaud et~al.(2015)Duvenaud, Maclaurin, Iparraguirre, Bombarell,
  Hirzel, Aspuru-Guzik, and Adams]{duvenaud2015convolutional}
Duvenaud, D.~K., Maclaurin, D., Iparraguirre, J., Bombarell, R., Hirzel, T.,
  Aspuru-Guzik, A., and Adams, R.~P.
\newblock Convolutional networks on graphs for learning molecular fingerprints.
\newblock In \emph{Advances in neural information processing systems}, pp.\
  2224--2232, 2015.

\bibitem[Dwivedi et~al.(2020)Dwivedi, Joshi, Laurent, Bengio, and
  Bresson]{dwivedi2020benchmarking}
Dwivedi, V.~P., Joshi, C.~K., Laurent, T., Bengio, Y., and Bresson, X.
\newblock Benchmarking graph neural networks.
\newblock \emph{arXiv preprint arXiv:2003.00982}, 2020.

\bibitem[Geng et~al.(2019)Geng, Li, Wang, Zhang, Yang, Ye, and
  Liu]{geng2019spatiotemporal}
Geng, X., Li, Y., Wang, L., Zhang, L., Yang, Q., Ye, J., and Liu, Y.
\newblock Spatiotemporal multi-graph convolution network for ride-hailing
  demand forecasting.
\newblock In \emph{Proceedings of the AAAI conference on artificial
  intelligence}, volume~33, pp.\  3656--3663, 2019.

\bibitem[Gilmer et~al.(2017)Gilmer, Schoenholz, Riley, Vinyals, and
  Dahl]{gilmer2017neural}
Gilmer, J., Schoenholz, S.~S., Riley, P.~F., Vinyals, O., and Dahl, G.~E.
\newblock Neural message passing for quantum chemistry.
\newblock In \emph{Proceedings of the 34th International Conference on Machine
  Learning-Volume 70}, pp.\  1263--1272. JMLR. org, 2017.

\bibitem[Guo et~al.(2022)Guo, Zhou, Hu, Li, Chang, and Wang]{guo2022orthogonal}
Guo, K., Zhou, K., Hu, X., Li, Y., Chang, Y., and Wang, X.
\newblock Orthogonal graph neural networks.
\newblock In \emph{Proceedings of the AAAI Conference on Artificial
  Intelligence}, volume~36, pp.\  3996--4004, 2022.

\bibitem[Hamilton et~al.(2017)Hamilton, Ying, and
  Leskovec]{hamilton2017inductive}
Hamilton, W., Ying, Z., and Leskovec, J.
\newblock Inductive representation learning on large graphs.
\newblock In \emph{Advances in Neural Information Processing Systems}, pp.\
  1024--1034, 2017.

\bibitem[Hammond et~al.(2011)Hammond, Vandergheynst, and
  Gribonval]{hammond2011wavelets}
Hammond, D.~K., Vandergheynst, P., and Gribonval, R.
\newblock Wavelets on graphs via spectral graph theory.
\newblock \emph{Applied and Computational Harmonic Analysis}, 30\penalty0
  (2):\penalty0 129--150, 2011.

\bibitem[He et~al.(2021)He, Wei, Huang, and Xu]{he2021bernnet}
He, M., Wei, Z., Huang, Z., and Xu, H.
\newblock Bernnet: Learning arbitrary graph spectral filters via bernstein
  approximation.
\newblock In \emph{NeurIPS}, 2021.

\bibitem[He et~al.(2022)He, Hooi, Laurent, Perold, LeCun, and
  Bresson]{he2022generalization}
He, X., Hooi, B., Laurent, T., Perold, A., LeCun, Y., and Bresson, X.
\newblock A generalization of vit/mlp-mixer to graphs, 2022.

\bibitem[Hornik et~al.(1989)Hornik, Stinchcombe, and
  White]{hornik1989multilayer}
Hornik, K., Stinchcombe, M., and White, H.
\newblock Multilayer feedforward networks are universal approximators.
\newblock \emph{Neural networks}, 2\penalty0 (5):\penalty0 359--366, 1989.

\bibitem[Huang et~al.(2020)Huang, Rong, Xu, Sun, and Huang]{huang2020tackling}
Huang, W., Rong, Y., Xu, T., Sun, F., and Huang, J.
\newblock Tackling over-smoothing for general graph convolutional networks.
\newblock \emph{arXiv preprint arXiv:2008.09864}, 2020.

\bibitem[Ivanov \& Burnaev(2018)Ivanov and Burnaev]{pmlr-v80-ivanov18a}
Ivanov, S. and Burnaev, E.
\newblock Anonymous walk embeddings.
\newblock In Dy, J. and Krause, A. (eds.), \emph{Proceedings of the 35th
  International Conference on Machine Learning}, volume~80 of \emph{Proceedings
  of Machine Learning Research}, pp.\  2191--2200, Stockholmsmässan, Stockholm
  Sweden, 10--15 Jul 2018. PMLR.
\newblock URL \url{http://proceedings.mlr.press/v80/ivanov18a.html}.

\bibitem[Jiang et~al.(2022)Jiang, Yang, Wen, Su, and Huang]{Jiang2022ASE}
Jiang, X., Yang, Z., Wen, P., Su, L., and Huang, Q.
\newblock A sparse-motif ensemble graph convolutional network against
  over-smoothing.
\newblock In \emph{IJCAI}, 2022.

\bibitem[Jin et~al.(2022)Jin, Liu, Ma, Aggarwal, and Tang]{jin2022feature}
Jin, W., Liu, X., Ma, Y., Aggarwal, C., and Tang, J.
\newblock Feature overcorrelation in deep graph neural networks: A new
  perspective.
\newblock In \emph{Proceedings of the 28th {ACM} {SIGKDD} Conference on
  Knowledge Discovery and Data Mining}, 2022.

\bibitem[Khan \& Blumenstock(2019)Khan and Blumenstock]{khan2019multi}
Khan, M.~R. and Blumenstock, J.~E.
\newblock Multi-gcn: Graph convolutional networks for multi-view networks, with
  applications to global poverty.
\newblock In \emph{Proceedings of the AAAI conference on artificial
  intelligence}, volume~33, pp.\  606--613, 2019.

\bibitem[Kipf \& Welling(2017)Kipf and Welling]{kipf2017semi}
Kipf, T.~N. and Welling, M.
\newblock Semi-supervised classification with graph convolutional networks.
\newblock In \emph{International Conference on Learning Representations
  (ICLR)}, 2017.

\bibitem[Klicpera et~al.(2019{\natexlab{a}})Klicpera, Bojchevski, and
  G{\"u}nnemann]{klicpera_predict_2019}
Klicpera, J., Bojchevski, A., and G{\"u}nnemann, S.
\newblock Predict then propagate: Graph neural networks meet personalized
  pagerank.
\newblock In \emph{International Conference on Learning Representations
  (ICLR)}, 2019{\natexlab{a}}.

\bibitem[Klicpera et~al.(2019{\natexlab{b}})Klicpera, Wei{\ss}enberger, and
  G{\"u}nnemann]{klicpera2019diffusion}
Klicpera, J., Wei{\ss}enberger, S., and G{\"u}nnemann, S.
\newblock Diffusion improves graph learning.
\newblock \emph{Advances in Neural Information Processing Systems},
  32:\penalty0 13354--13366, 2019{\natexlab{b}}.

\bibitem[Kreuzer et~al.(2021)Kreuzer, Beaini, Hamilton, Létourneau, and
  Tossou]{kreuzer2021rethinking}
Kreuzer, D., Beaini, D., Hamilton, W., Létourneau, V., and Tossou, P.
\newblock Rethinking graph transformers with spectral attention.
\newblock \emph{arXiv preprint arXiv:2106.03893}, 2021.

\bibitem[Le et~al.(2021)Le, Bertolini, No{\'e}, and
  Clevert]{le2021parameterized}
Le, T., Bertolini, M., No{\'e}, F., and Clevert, D.-A.
\newblock Parameterized hypercomplex graph neural networks for graph
  classification.
\newblock \emph{arXiv preprint arXiv:2103.16584}, 2021.

\bibitem[Lee et~al.(2019)Lee, Rossi, Kong, Kim, Koh, and Rao]{lee2019graph}
Lee, J.~B., Rossi, R.~A., Kong, X., Kim, S., Koh, E., and Rao, A.
\newblock Graph convolutional networks with motif-based attention.
\newblock In \emph{Proceedings of the 28th ACM international conference on
  information and knowledge management}, pp.\  499--508, 2019.

\bibitem[Levie et~al.(2019)Levie, Monti, Bresson, and
  Bronstein]{ron2019cayleynets}
Levie, R., Monti, F., Bresson, X., and Bronstein, M.~M.
\newblock Cayleynets: Graph convolutional neural networks with complex rational
  spectral filters.
\newblock \emph{IEEE Transactions on Signal Processing}, 67\penalty0
  (1):\penalty0 97--109, 2019.
\newblock \doi{10.1109/TSP.2018.2879624}.

\bibitem[Li et~al.(2020)Li, Xiong, Thabet, and Ghanem]{li2020deepergcn}
Li, G., Xiong, C., Thabet, A., and Ghanem, B.
\newblock Deepergcn: All you need to train deeper gcns.
\newblock \emph{arXiv preprint arXiv:2006.07739}, 2020.

\bibitem[Li et~al.(2018)Li, Han, and Wu]{li2018deeper}
Li, Q., Han, Z., and Wu, X.-M.
\newblock Deeper insights into graph convolutional networks for semi-supervised
  learning.
\newblock In \emph{Thirty-Second AAAI Conference on Artificial Intelligence},
  2018.

\bibitem[Liu et~al.(2020)Liu, Gao, and Ji]{liu2020towards}
Liu, M., Gao, H., and Ji, S.
\newblock Towards deeper graph neural networks.
\newblock In \emph{Proceedings of the 26th ACM SIGKDD International Conference
  on Knowledge Discovery \& Data Mining}, pp.\  338--348, 2020.

\bibitem[Maron et~al.(2018)Maron, Ben-Hamu, Shamir, and
  Lipman]{maron2018invariant}
Maron, H., Ben-Hamu, H., Shamir, N., and Lipman, Y.
\newblock Invariant and equivariant graph networks.
\newblock \emph{arXiv preprint arXiv:1812.09902}, 2018.

\bibitem[Maron et~al.(2019)Maron, Ben-Hamu, Serviansky, and
  Lipman]{maron2019provably}
Maron, H., Ben-Hamu, H., Serviansky, H., and Lipman, Y.
\newblock Provably powerful graph networks.
\newblock In \emph{Proceedings of the 33rd International Conference on Neural
  Information Processing Systems}, pp.\  2156--2167, 2019.

\bibitem[Min et~al.(2022)Min, Chen, Bian, Xu, Zhao, Huang, Zhao, Huang,
  Ananiadou, and Rong]{min2022transformer}
Min, E., Chen, R., Bian, Y., Xu, T., Zhao, K., Huang, W., Zhao, P., Huang, J.,
  Ananiadou, S., and Rong, Y.
\newblock Transformer for graphs: An overview from architecture perspective.
\newblock \emph{arXiv preprint arXiv:2202.08455}, 2022.

\bibitem[Ming~Chen et~al.(2020)Ming~Chen, Zengfeng~Huang, and
  Li]{chenWHDL2020gcnii}
Ming~Chen, Z.~W., Zengfeng~Huang, B.~D., and Li, Y.
\newblock Simple and deep graph convolutional networks.
\newblock 2020.

\bibitem[Morris et~al.(2019)Morris, Ritzert, Fey, Hamilton, Lenssen, Rattan,
  and Grohe]{morris2019weisfeiler}
Morris, C., Ritzert, M., Fey, M., Hamilton, W.~L., Lenssen, J.~E., Rattan, G.,
  and Grohe, M.
\newblock Weisfeiler and leman go neural: Higher-order graph neural networks.
\newblock In \emph{Proceedings of the AAAI Conference on Artificial
  Intelligence}, volume~33, pp.\  4602--4609, 2019.

\bibitem[Neumann et~al.(2016)Neumann, Garnett, Bauckhage, and
  Kersting]{neumann2016propagation}
Neumann, M., Garnett, R., Bauckhage, C., and Kersting, K.
\newblock Propagation kernels: efficient graph kernels from propagated
  information.
\newblock \emph{Machine Learning}, 102\penalty0 (2):\penalty0 209--245, 2016.

\bibitem[Niepert et~al.(2016)Niepert, Ahmed, and Kutzkov]{niepert2016learning}
Niepert, M., Ahmed, M., and Kutzkov, K.
\newblock Learning convolutional neural networks for graphs.
\newblock In \emph{International Conference on Machine Learning}, pp.\
  2014--2023, 2016.

\bibitem[Oono \& Suzuki(2020)Oono and Suzuki]{oono2020graph}
Oono, K. and Suzuki, T.
\newblock Graph neural networks exponentially lose expressive power for node
  classification.
\newblock In \emph{International Conference on Learning Representations}, 2020.
\newblock URL \url{https://openreview.net/forum?id=S1ldO2EFPr}.

\bibitem[Ortega et~al.(2018)Ortega, Frossard, Kova{\v{c}}evi{\'c}, Moura, and
  Vandergheynst]{ortega2018graph}
Ortega, A., Frossard, P., Kova{\v{c}}evi{\'c}, J., Moura, J.~M., and
  Vandergheynst, P.
\newblock Graph signal processing: Overview, challenges, and applications.
\newblock \emph{Proceedings of the IEEE}, 106\penalty0 (5):\penalty0 808--828,
  2018.

\bibitem[Ramp\'{a}\v{s}ek et~al.(2022)Ramp\'{a}\v{s}ek, Galkin, Dwivedi, Luu,
  Wolf, and Beaini]{rampasek2022GPS}
Ramp\'{a}\v{s}ek, L., Galkin, M., Dwivedi, V.~P., Luu, A.~T., Wolf, G., and
  Beaini, D.
\newblock {Recipe for a General, Powerful, Scalable Graph Transformer}.
\newblock \emph{arXiv:2205.12454}, 2022.

\bibitem[Rong et~al.(2020)Rong, Huang, Xu, and Huang]{Rong2020DropEdge:}
Rong, Y., Huang, W., Xu, T., and Huang, J.
\newblock Dropedge: Towards deep graph convolutional networks on node
  classification.
\newblock In \emph{International Conference on Learning Representations}, 2020.
\newblock URL \url{https://openreview.net/forum?id=Hkx1qkrKPr}.

\bibitem[Sandryhaila \& Moura(2013)Sandryhaila and
  Moura]{sandryhaila2013discrete}
Sandryhaila, A. and Moura, J.~M.
\newblock Discrete signal processing on graphs.
\newblock \emph{IEEE transactions on signal processing}, 61\penalty0
  (7):\penalty0 1644--1656, 2013.

\bibitem[Sato(2020)]{sato2020survey}
Sato, R.
\newblock A survey on the expressive power of graph neural networks.
\newblock \emph{arXiv preprint arXiv:2003.04078}, 2020.

\bibitem[Shervashidze et~al.(2009)Shervashidze, Vishwanathan, Petri, Mehlhorn,
  and Borgwardt]{shervashidze2009efficient}
Shervashidze, N., Vishwanathan, S., Petri, T., Mehlhorn, K., and Borgwardt, K.
\newblock Efficient graphlet kernels for large graph comparison.
\newblock In \emph{Artificial Intelligence and Statistics}, pp.\  488--495,
  2009.

\bibitem[Shuman et~al.(2013)Shuman, Narang, Frossard, Ortega, and
  Vandergheynst]{shuman2013emerging}
Shuman, D.~I., Narang, S.~K., Frossard, P., Ortega, A., and Vandergheynst, P.
\newblock The emerging field of signal processing on graphs: Extending
  high-dimensional data analysis to networks and other irregular domains.
\newblock \emph{IEEE signal processing magazine}, 30\penalty0 (3):\penalty0
  83--98, 2013.

\bibitem[Simonovsky \& Komodakis(2017)Simonovsky and
  Komodakis]{simonovsky2017dynamic}
Simonovsky, M. and Komodakis, N.
\newblock Dynamic edge-conditioned filters in convolutional neural networks on
  graphs.
\newblock In \emph{Proceedings of the IEEE conference on computer vision and
  pattern recognition}, pp.\  3693--3702, 2017.

\bibitem[Veli{\v{c}}kovi{\'{c}} et~al.(2018)Veli{\v{c}}kovi{\'{c}}, Cucurull,
  Casanova, Romero, Li{\`{o}}, and Bengio]{velickovic2018graph}
Veli{\v{c}}kovi{\'{c}}, P., Cucurull, G., Casanova, A., Romero, A., Li{\`{o}},
  P., and Bengio, Y.
\newblock {Graph Attention Networks}.
\newblock \emph{International Conference on Learning Representations}, 2018.
\newblock URL \url{https://openreview.net/forum?id=rJXMpikCZ}.

\bibitem[Verma \& Zhang(2017)Verma and Zhang]{verma2017hunt}
Verma, S. and Zhang, Z.-L.
\newblock Hunt for the unique, stable, sparse and fast feature learning on
  graphs.
\newblock In \emph{Advances in Neural Information Processing Systems}, pp.\
  88--98, 2017.

\bibitem[Vishwanathan et~al.(2010)Vishwanathan, Schraudolph, Kondor, and
  Borgwardt]{vishwanathan2010graph}
Vishwanathan, S. V.~N., Schraudolph, N.~N., Kondor, R., and Borgwardt, K.~M.
\newblock Graph kernels.
\newblock \emph{Journal of Machine Learning Research}, 11\penalty0
  (Apr):\penalty0 1201--1242, 2010.

\bibitem[Wang \& Zhang(2022)Wang and Zhang]{JacobiConv}
Wang, X. and Zhang, M.
\newblock How powerful are spectral graph neural networks.
\newblock \emph{ICML}, 2022.

\bibitem[Wu et~al.(2019{\natexlab{a}})Wu, Souza, Zhang, Fifty, Yu, and
  Weinberger]{pmlr-v97-wu19e}
Wu, F., Souza, A., Zhang, T., Fifty, C., Yu, T., and Weinberger, K.
\newblock Simplifying graph convolutional networks.
\newblock In \emph{Proceedings of the 36th International Conference on Machine
  Learning}, pp.\  6861--6871. PMLR, 2019{\natexlab{a}}.

\bibitem[Wu et~al.(2020)Wu, Pan, Zhou, Chang, and Zhu]{wu2020unsupervised}
Wu, M., Pan, S., Zhou, C., Chang, X., and Zhu, X.
\newblock Unsupervised domain adaptive graph convolutional networks.
\newblock In \emph{Proceedings of The Web Conference 2020}, pp.\  1457--1467,
  2020.

\bibitem[Wu et~al.(2019{\natexlab{b}})Wu, Pan, Chen, Long, Zhang, and
  Yu]{wu2019comprehensive}
Wu, Z., Pan, S., Chen, F., Long, G., Zhang, C., and Yu, P.~S.
\newblock A comprehensive survey on graph neural networks.
\newblock \emph{arXiv preprint arXiv:1901.00596}, 2019{\natexlab{b}}.

\bibitem[Xanthopoulos et~al.(2013)Xanthopoulos, Pardalos, Trafalis,
  Xanthopoulos, Pardalos, and Trafalis]{xanthopoulos2013linear}
Xanthopoulos, P., Pardalos, P.~M., Trafalis, T.~B., Xanthopoulos, P., Pardalos,
  P.~M., and Trafalis, T.~B.
\newblock Linear discriminant analysis.
\newblock \emph{Robust data mining}, pp.\  27--33, 2013.

\bibitem[Xinyi \& Chen(2019)Xinyi and Chen]{xinyi2018capsule}
Xinyi, Z. and Chen, L.
\newblock Capsule graph neural network.
\newblock In \emph{International Conference on Learning Representations}, 2019.
\newblock URL \url{https://openreview.net/forum?id=Byl8BnRcYm}.

\bibitem[Xu et~al.(2019{\natexlab{a}})Xu, Shen, Cao, Qiu, and
  Cheng]{xu2018graph}
Xu, B., Shen, H., Cao, Q., Qiu, Y., and Cheng, X.
\newblock Graph wavelet neural network.
\newblock In \emph{International Conference on Learning Representations},
  2019{\natexlab{a}}.
\newblock URL \url{https://openreview.net/forum?id=H1ewdiR5tQ}.

\bibitem[Xu et~al.(2018)Xu, Li, Tian, Sonobe, Kawarabayashi, and
  Jegelka]{xu2018representation}
Xu, K., Li, C., Tian, Y., Sonobe, T., Kawarabayashi, K.-i., and Jegelka, S.
\newblock Representation learning on graphs with jumping knowledge networks.
\newblock In \emph{International Conference on Machine Learning}, pp.\
  5453--5462. PMLR, 2018.

\bibitem[Xu et~al.(2019{\natexlab{b}})Xu, Hu, Leskovec, and Jegelka]{xu2018how}
Xu, K., Hu, W., Leskovec, J., and Jegelka, S.
\newblock How powerful are graph neural networks?
\newblock In \emph{International Conference on Learning Representations},
  2019{\natexlab{b}}.
\newblock URL \url{https://openreview.net/forum?id=ryGs6iA5Km}.

\bibitem[Yanardag \& Vishwanathan(2015)Yanardag and
  Vishwanathan]{yanardag2015deep}
Yanardag, P. and Vishwanathan, S.
\newblock Deep graph kernels.
\newblock In \emph{Proceedings of the 21th ACM SIGKDD International Conference
  on Knowledge Discovery and Data Mining}, pp.\  1365--1374. ACM, 2015.

\bibitem[Yang et~al.(2020)Yang, Shen, Qi, and Yin]{yang2020breaking}
Yang, M., Shen, Y., Qi, H., and Yin, B.
\newblock Breaking the expressive bottlenecks of graph neural networks.
\newblock \emph{arXiv preprint arXiv:2012.07219}, 2020.

\bibitem[Yang et~al.(2022{\natexlab{a}})Yang, Shen, Li, Qi, Zhang, and
  Yin]{yang2022spectrum}
Yang, M., Shen, Y., Li, R., Qi, H., Zhang, Q., and Yin, B.
\newblock A new perspective on the effects of spectrum in graph neural
  networks.
\newblock In \emph{Proceedings of the 39th International Conference on Machine
  Learning}, 2022{\natexlab{a}}.

\bibitem[Yang et~al.(2022{\natexlab{b}})Yang, Wang, Shen, Qi, and
  Yin]{yang2022expc}
Yang, M., Wang, R., Shen, Y., Qi, H., and Yin, B.
\newblock Breaking the expression bottleneck of graph neural networks.
\newblock \emph{IEEE Transactions on Knowledge and Data Engineering}, pp.\
  1--1, 2022{\natexlab{b}}.
\newblock \doi{10.1109/TKDE.2022.3168070}.

\bibitem[Ying et~al.(2021)Ying, Cai, Luo, Zheng, Ke, He, Shen, and
  Liu]{ying2021transformers}
Ying, C., Cai, T., Luo, S., Zheng, S., Ke, G., He, D., Shen, Y., and Liu, T.-Y.
\newblock Do transformers really perform bad for graph representation?
\newblock \emph{arXiv preprint arXiv:2106.05234}, 2021.

\bibitem[Ying et~al.(2018)Ying, You, Morris, Ren, Hamilton, and
  Leskovec]{ying2018hierarchical}
Ying, Z., You, J., Morris, C., Ren, X., Hamilton, W., and Leskovec, J.
\newblock Hierarchical graph representation learning with differentiable
  pooling.
\newblock In \emph{Advances in Neural Information Processing Systems}, pp.\
  4800--4810, 2018.

\bibitem[Zhang et~al.(2018)Zhang, Cui, Neumann, and Chen]{zhang2018end}
Zhang, M., Cui, Z., Neumann, M., and Chen, Y.
\newblock An end-to-end deep learning architecture for graph classification.
\newblock In \emph{Thirty-Second AAAI Conference on Artificial Intelligence},
  2018.

\bibitem[Zhao \& Akoglu(2020)Zhao and Akoglu]{zhao2020pairnorm}
Zhao, L. and Akoglu, L.
\newblock Pairnorm: Tackling oversmoothing in gnns.
\newblock In \emph{International Conference on Learning Representations}, 2020.
\newblock URL \url{https://openreview.net/forum?id=rkecl1rtwB}.

\bibitem[Zhou \& Sch{\"o}lkopf(2004)Zhou and
  Sch{\"o}lkopf]{zhou2004regularization}
Zhou, D. and Sch{\"o}lkopf, B.
\newblock A regularization framework for learning from graph data.
\newblock In \emph{ICML 2004 Workshop on Statistical Relational Learning and
  Its Connections to Other Fields (SRL 2004)}, pp.\  132--137, 2004.

\bibitem[Zhou et~al.(2021)Zhou, Dong, Wang, Lee, Hooi, Xu, and
  Feng]{zhou2021understanding}
Zhou, K., Dong, Y., Wang, K., Lee, W.~S., Hooi, B., Xu, H., and Feng, J.
\newblock Understanding and resolving performance degradation in deep graph
  convolutional networks.
\newblock In \emph{Proceedings of the 30th ACM International Conference on
  Information \& Knowledge Management}, pp.\  2728--2737, 2021.

\bibitem[Zhu \& Koniusz(2020)Zhu and Koniusz]{zhu2020simple}
Zhu, H. and Koniusz, P.
\newblock Simple spectral graph convolution.
\newblock In \emph{International Conference on Learning Representations}, 2020.

\end{thebibliography}

	\newpage
	\appendix
	\onecolumn
	

	\section{Graph Representation.}
	\label{appe:graph_representation}
	
	Considering the various possible representations for the topology of a graph,
	we denote the matrix representation space of an undirected graph $G$ by $\gM_{G}$.
	Admittedly, providing the formal unified definition for $\gM_{G}$ or enumerating all its elements is indeed hard,
	but it is still possible to give some feasible instances.
	For example, $\gM_{G}$ can include
	\begin{compactitem}
		\item Graph shift operators~\citep{sandryhaila2013discrete}: including the adjacency matrix, Laplacian matrix, and their various normalization versions, as well as the mean aggregation operator of GNNs and Parametrized graph shift operator (PGSO)~\citep{dasoulas2021learning};
		\item Structure derived matrices: $k$ path-length counting matrix $\rmA^{k}$, shortest path distance matrix (SPD),  motif adjacency matrix~\citep{lee2019graph,Jiang2022ASE}, and point-wise mutual information matrix~\citep{wu2020unsupervised}, etc.
		\item Feature-engineering-based matrices, like the neural graph fingerprint~\citep{duvenaud2015convolutional}.
	\end{compactitem}
	Different matrices emphasize the graph structure or topology from different angles, like local or global view;
	and each of them has its own limitations in that there are some properties that the matrix cannot always determine~\citep{butler2017spectral}.
	Under the graph spectral formulation, here we only account for $\gM_{G}$ which is composed of some symmetric matrices;
	for any $S \in \gM_{G}$, it has the unique eigendecomposition as $S = \rmU \rmLambda \rmU^{\top}$, according to the spectral theorem.

	\section{Proofs.}
	\subsection{Proof of Proposition~\ref{prop:smoothness}}
	\label{proof:prop:smoothness}
	\begin{proof}
		Let $\vf^{\prime}=g_{\vtheta}(\rmL) \vf$ denote the graph signal after graph convolution.
		If $|g_{\vtheta}(\vlambda_{i})|<1, i\in[n]$, then
		\begin{equation}
			\nonumber
			\begin{aligned}
				\vf^{\prime\top} \rmL \vf^{\prime}&=(g_{\vtheta}(\rmL) \vf)^{\top} \rmL g_{\vtheta}(\rmL) \vf\\
				&=\vf^{\top} g_{\vtheta}(\rmL) \rmL g_{\vtheta}(\rmL) \vf\\
				&=\vf^{\top} \rmU g_{\vtheta}(\rmLambda) \rmLambda g_{\vtheta}(\rmLambda) \rmU^{\top} \vf\\
				&=\vf^{\top} \rmU \mathrm{diag}(g_{\vtheta}(\vlambda_{i})^2 \vlambda_{i}) \rmU^{\top} \vf\\
				&=\hat {\vf}^{\top} \mathrm{diag}(g_{\vtheta}(\vlambda_{i})^2 \vlambda_{i}) \hat {\vf}\\
				&=\sum_{i=1}^{n} \hat {\vf}_{i}^{2}\vlambda_{i} g_{\vtheta}(\vlambda_{i})^2\\
				&<\sum_{i=1}^{n} \hat {\vf}_{i}^{2}\vlambda_{i} \qquad\quad/* \vlambda_{i}\geq 0, i\in[n] */\\
				&=\vf^{\top} \rmL \vf \qquad\qquad/*\textrm{As in Eq.~\ref{equ:weighted_inner_product}}*/
			\end{aligned}
		\end{equation}
		Hence, $\vf^{\prime}$ is smoother than $\vf$ w.r.t. $\rmL$.
		Similarly, we can prove $\vf^{\prime\top} \rmL \vf^{\prime} > \vf^{\top} \rmL \vf$ $\vf^{\prime}$ when $|g_{\vtheta}(\vlambda_{i})|>1, i\in[n]$.
	\end{proof}
	
	\subsection{Proof of Proposition~\ref{prop:gcn_smoothness}}
	\label{proof:prop:gcn_smoothness}
	\begin{proof}
		In GCN, we have $\vf^{\prime}=\tilde \rmD^{-\frac{1}{2}}\tilde \rmA\tilde \rmD^{-\frac{1}{2}}\vf$.
		Let $\tilde \rmL=\rmI-\tilde \rmD^{-\frac{1}{2}}\tilde \rmA\tilde \rmD^{-\frac{1}{2}}$.
		Then $\tilde \rmD^{-\frac{1}{2}}\tilde \rmA\tilde \rmD^{-\frac{1}{2}}=\rmI-\tilde \rmL=g_{\vtheta}(\tilde \rmL)$ is the polynomial of the Laplacian $\tilde \rmL$.
		Let $\tilde{\vlambda}_{i}, i\in[n]$ denote the spectrum of $\tilde \rmL$.
		Then, $0\leq\tilde{\vlambda}_{i}<2$ and $g_{\vtheta}(\tilde{\vlambda}_{i})=1-\tilde{\vlambda}_{i}\in(-1, 1)$.
		According to Proposition~\ref{prop:smoothness}, we have $\vf^{\prime\top}\tilde \rmL\vf^{\prime}\leq\vf^{\top}\tilde \rmL\vf$,
	\end{proof}

	\section{Experimental Details.}
	
	\subsection{Datasets Statistics.}
	\label{appe:dataset_statistics}
	
	All detailed statistics of the datasets used in our experiments are presented in Tab.~\ref{tab:dataset_statistics}.
	The corresponding tasks involve graph regression task and graph classification task collecting from real-world molecules, social networks and protein-protein interactions.
	The scale of datasets ranges from hundreds of graphs (e.g. MUTAG, PTC\_MR) to hundreds of thousands of graphs (e.g., ogbg-molpcba, pgbg-ppa).
	The scale of graphs involved in each dataset ranges from 10-20 (e.g., MUTAG, PTC\_MR, IMDB-B) to 400-500 (e.g., RDT-B).
	Also, the density of connectivity e.g., $\frac{2 \times \textrm{Avg \# edges}}{\textrm{Avg \# nodes}}$ ranges from 2.x (most molecular datasets) to 18.x (e.g., ogbg-ppa).
	\begin{table}[h]
		\caption{Statistics of the datasets used in our experiments.}
		\label{tab:dataset_statistics}
		\centering
		\vspace{5pt}
			\begin{tabular}{l|cccccc}\toprule
				\textbf{Dataset} &\textbf{\# Graphs} &\textbf{Avg \# nodes} &\textbf{Avg \# edges} &\textbf{Node attr} &\textbf{Edge attr} &\textbf{Task type}\\
				\midrule
				ZINC &12,000 &23.2 &24.9 &Y &Y &Regression\\
				\midrule
				ogbg-molpcba &437,929 &26.0 &28.1 &Y &Y &Binary classi.\\
				ogbg-ppa &158,100 &243.4 &2,266.1 &N &Y &37-way classi.\\
				\midrule
				MUTAG &188 &17.93 &19.79 &N &N &Binary classi.\\
				NCI1 &4110 &29.87 &32.39 &N &N &Binary classi.\\
				NCI109 &4127 &29.68 &32.13 &N &N &Binary classi.\\
				ENZYMES &600 &32.63 &62.14 &Y &N &6-way classi\\
				PTC\_MR &344 &14.29 &14.69 &N &N &Binary classi.\\
				PROTEINS &1113 &39.06 &72.82 &Y &N &Binary classi.\\
				IMDB-B &1000 &19.77 &96.53 &N &N &Binary classi.\\
				RDT-B &2000 &429.63 &497.75 &N &N &Binary classi.\\
				\bottomrule
			\end{tabular}
		\vspace{-10pt}
	\end{table}
	
	\subsection{Baselines.}
	\label{appe:baseline}
	The baseline models used for comparisons include: 
	GK~\citep{shervashidze2009efficient},
	RW~\citep{vishwanathan2010graph},
	PK~\citep{neumann2016propagation},
	FGSD~\citep{verma2017hunt},
	AWE~\citep{pmlr-v80-ivanov18a},
	DGCNN~\citep{zhang2018end},
	PSCN~\citep{niepert2016learning},
	DCNN~\citep{atwood2016diffusion},
	ECC~\citep{simonovsky2017dynamic},
	DGK~\citep{yanardag2015deep},
	CapsGNN~\citep{xinyi2018capsule},
	DiffPool~\citep{ying2018hierarchical},
	GIN~\citep{xu2018how},
	$k$-GNN~\citep{morris2019weisfeiler},
	IGN~\citep{maron2018invariant},
	PPGNN~\citep{maron2019provably},
	GCN$^2$~\citep{de2020natural}
	GraphSage~\citep{hamilton2017inductive},
	GAT~\citep{velickovic2018graph},
	GatedGCN-PE~\citep{bresson2017residual},
	MPNN (sum)~\citep{gilmer2017neural},
	DeeperG~\citep{li2020deepergcn},
	PNA~\citep{corso2020pna},
	DGN~\citep{beani2021directional},
	GSN~\citep{bouritsas2020improving},
	GINE-{\scriptsize APPNP}~\citep{brossard2020graph},
	PHC-GNN~\citep{le2021parameterized},
	ExpC~\citep{yang2022expc},
	GT~\citep{dwivedi2020benchmarking},
	SAN~\citep{kreuzer2021rethinking},
	Graphormer~\citep{ying2021transformers},
	KS-SAT~\citep{chen2022structure},
	GPS~\citep{rampasek2022GPS},
	GM-Mix~\citep{he2022generalization}.
	
	\subsection{Experimental Setup.}
	\label{appe:setup}
	
	Tab.~\ref{tab:zinc_ogbg_settings} and Tab.~\ref{tab:tu_settings} present all hyperparameter configurations used in baseline comparisons in Sec.~\ref{sec:results}.
	On ZINC, we keep the number of learnable coefficients used on our model close to 500K as configured by other baseline methods.
	\begin{table}[h]
		\centering
		\caption{Hyperparameter settings on ZINC and OGB datasets.}
		\label{tab:zinc_ogbg_settings}
		\vspace{5pt}
			\begin{tabular}{c|ccc}
				\toprule
				Hyperparameter &ZINC &ogbg-molpcba &ogbg-ppa\\
				\midrule
				Hidden Dim. &160 &384 &384\\
				Num. Layers &6 &8 &4\\
				Drop. Rate &0 &0 &0.5\\
				Readout &Mean &Max &Sum\\
				\midrule
				Batch Size &64 &64 &16\\
				Initial LR &0.001 &0.0005 &0.001\\
				LR Dec. Steps &35 &5 &30\\
				LR Dec. Rate &0.6 &0.2 &0.65\\
				\# Warm. Steps &5 &5 &5\\
				Weight Dec. &5e-5 &1e-2 &0\\
				\midrule
				$\gG$ &Dense &Dense &Sparse\\
				$\sigma$ in $f_{\vtheta}$ &GELU &GELU &GELU\\
				\midrule
				$\{(\eps, k)\}$ &
				\makecell{(-0.1, 4),\\(-0.2, 4),\\(-0.3, 4),\\(-0.4, 4),\\(-0.5, 4)} &
				\makecell{(-0.2, 1),\\(-0.2, 2),\\(-0.2, 3),\\(-0.2, 4),\\(-0.2, 5),\\(-0.25, 1),\\(-0.25, 2),\\(-0.25, 3),\\(-0.25, 4),\\(-0.25, 5),\\(-0.3, 1),\\(-0.3, 2),\\(-0.3, 3),\\(-0.3, 4),\\(-0.3, 5),\\(-0.35, 1),\\(-0.35, 2),\\(-0.35, 3),\\(-0.35, 4),\\(-0.35, 5)} &
				\makecell{(0, 1),\\(-0.05, 1),\\(-0.1, 1),\\(-0.15, 1),\\(-0.2, 1),\\(-0.2, 2),\\(-0.25, 1),\\(-0.25, 2),\\(-0.3, 1),\\(-0.3, 2),\\(-0.35, 1),\\(-0.35, 2),\\(-0.4, 1),\\(-0.4, 2),\\(-0.4, 3),\\(-0.45, 1),\\(-0.45, 2),\\(-0.45, 3),\\(-0.5, 1),\\(-0.5, 2),\\(-0.5, 3)}\\
				\bottomrule
			\end{tabular}
		\vspace{-10pt}
	\end{table}
	
	\begin{table}[h]
		\centering
		\caption{Hyperparameter settings on TUDataset.}
		\label{tab:tu_settings}
		\vspace{5pt}
		\resizebox{1.0\textwidth}{!}{
			\begin{tabular}{c|cccccccc}
				\toprule
				Hyperparameter &MUTAG &NCI1 &NCI109 &ENZYMES &
				PTC\_MR &PROTEINS &IMDB-B &RDT-B\\
				\midrule
				Hidden Dim. &256 &256 &256 &256 &128 &128 &256 &256\\
				Num. Layers &4 &6 &6 &6 &6 &6 &3 &4\\
				Drop. Rate &0 &0 &0 &0.2 &0 &0 &0 &0\\
				Readout &Max &Max &Max &Max &Max &Mean &Max &Max\\
				\midrule
				Batch Size &64 &64 &64 &64 &64 &64 &16 &64\\
				Initial LR &0.001 &0.001 &0.001 &0.001 &0.001 &0.001 &0.001 &0.001\\
				LR Dec. Steps &50 &50 &50 &40 &30 &50 &40 &50\\
				LR Dec. Rate &0.6 &0.6 &0.6 &0.6 &0.65 &0.65 &0.6 &0.6\\
				\# Warm. Steps &0 &0 &0 &0 &0 &0 &0 &0\\
				Weight Dec. &0 &0 &0 &0 &0 &0 &0 &0\\
				\midrule
				$\gG$ &Dense &Dense &Dense &Dense &Dense &Dense &Dense &Sparse\\
				$\sigma$ in $f_{\vtheta}$ &GELU &GELU &GELU &GELU &GELU &GELU &GELU &GELU\\
				\midrule
				$\{(\eps, k)\}$ &
				\makecell{(-0.2, 1),\\(-0.2, 2),\\(-0.2, 3),\\(-0.2, 4),\\(-0.2, 5),\\(-0.25, 1),\\(-0.25, 2),\\(-0.25, 3),\\(-0.25, 4),\\(-0.25, 5),\\(-0.3, 1),\\(-0.3, 2),\\(-0.3, 3),\\(-0.3, 4),\\(-0.3, 5)} &
				\makecell{(-0.2, 1),\\(-0.2, 2),\\(-0.2, 3),\\(-0.2, 4),\\(-0.2, 5),\\(-0.25, 1),\\(-0.25, 2),\\(-0.25, 3),\\(-0.25, 4),\\(-0.25, 5)} &
				\makecell{(-0.2, 1),\\(-0.2, 2),\\(-0.2, 3),\\(-0.2, 4),\\(-0.2, 5),\\(-0.25, 1),\\(-0.25, 2),\\(-0.25, 3),\\(-0.25, 4),\\(-0.25, 5)} &
				\makecell{(-0.2, 1),\\(-0.2, 2),\\(-0.2, 3),\\(-0.2, 4),\\(-0.2, 5),\\(-0.2, 6),\\(-0.25, 1),\\(-0.25, 2),\\(-0.25, 3),\\(-0.25, 4),\\(-0.25, 5),\\(-0.25, 6),\\(-0.3, 1),\\(-0.3, 2),\\(-0.3, 3),\\(-0.3, 4),\\(-0.3, 5),\\(-0.3, 6),\\(-0.35, 1),\\(-0.35, 2),\\(-0.35, 3),\\(-0.35, 4),\\(-0.35, 5),\\(-0.35, 6)} &
				\makecell{(-0.3, 1),\\(-0.3, 2),\\(-0.3, 3),\\(-0.3, 4),\\(-0.3, 5),\\(-0.3, 6),\\(-0.3, 7),\\(-0.3, 8),\\(-0.35, 1),\\(-0.35, 2),\\(-0.35, 3),\\(-0.35, 4),\\(-0.35, 5),\\(-0.35, 6),\\(-0.35, 7),\\(-0.35, 8),\\(-0.4, 1),\\(-0.4, 2),\\(-0.4, 3),\\(-0.4, 4),\\(-0.4, 5),\\(-0.4, 6),\\(-0.4, 7),\\(-0.4, 8)} &
				\makecell{(-0.3, 1),\\(-0.3, 2),\\(-0.3, 3),\\(-0.3, 4),\\(-0.35, 1),\\(-0.35, 2),\\(-0.35, 3),\\(-0.35, 4),\\(-0.4, 1),\\(-0.4, 2),\\(-0.4, 3),\\(-0.4, 4)} &
				\makecell{(-0.2, 1),\\(-0.2, 2),\\(-0.2, 3),\\(-0.25, 1),\\(-0.25, 2),\\(-0.25, 3),\\(-0.3, 1),\\(-0.3, 2),\\(-0.3, 3),\\(-0.3, 4),\\(-0.3, 5),\\(-0.35, 1),\\(-0.35, 2),\\(-0.35, 3),\\(-0.35, 4),\\(-0.35, 5),\\(-0.4, 1),\\(-0.4, 2),\\(-0.4, 3),\\(-0.4, 4),\\(-0.4, 5)} &
				\makecell{(-0.3, 1),\\(-0.3, 2),\\(-0.35, 1),\\(-0.35, 2),\\(-0.4, 1),\\(-0.4, 2}\\
				\bottomrule
			\end{tabular}
		}
		\vspace{-10pt}
	\end{table}
	
	\subsection{Additional Experimental Details.}
	\label{appe:more_results}
	
\begin{figure*}[h]
		\centering
		\includegraphics[width=0.8\textwidth]{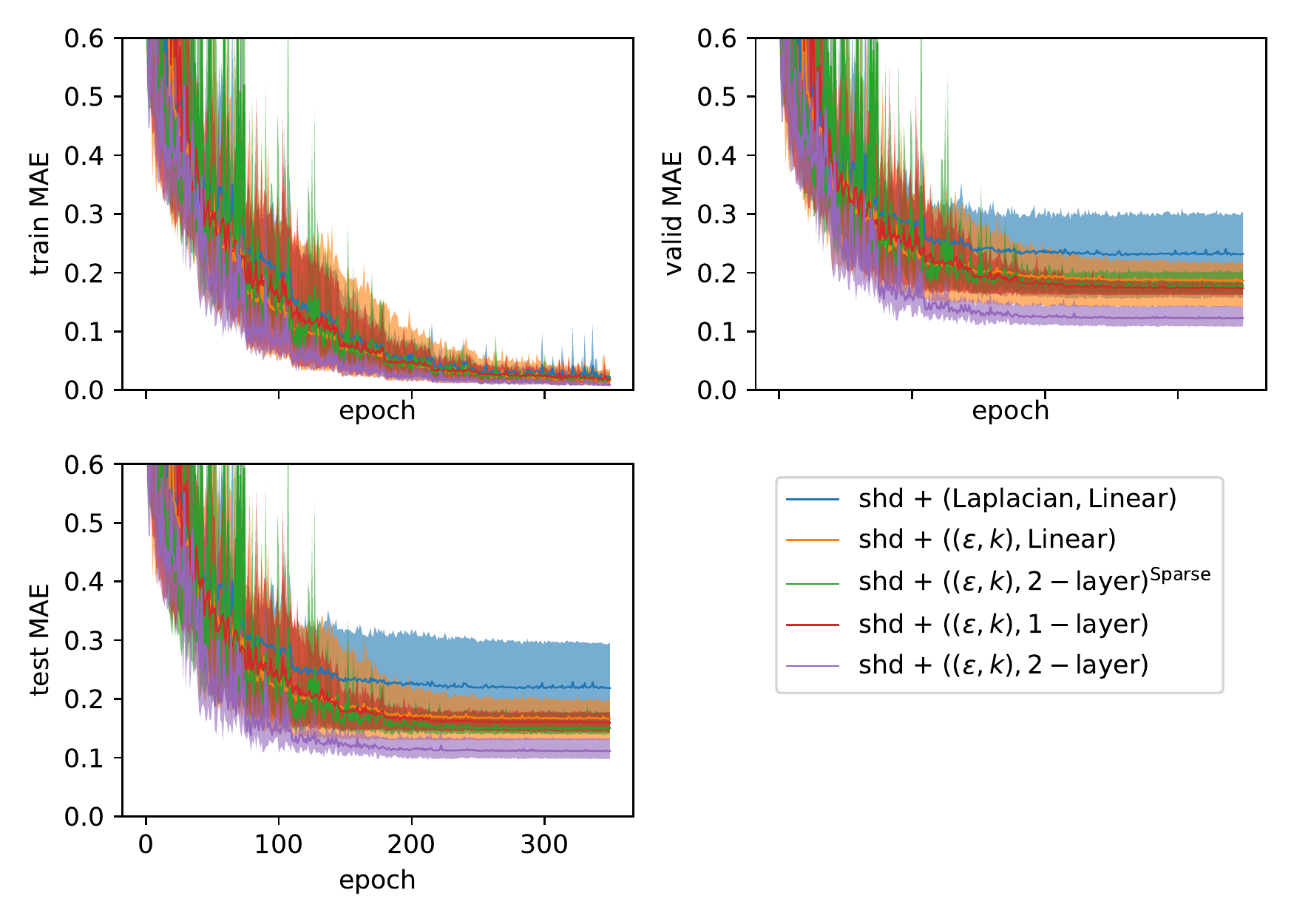}
		\vspace{-10pt}
		\caption{Ablation studies with channel-shared architecture on ZINC.}
		\label{fig:ablation_shd}
		\vspace{-10pt}
	\end{figure*}

\begin{figure*}[h]
		\centering
		\includegraphics[width=0.8\textwidth]{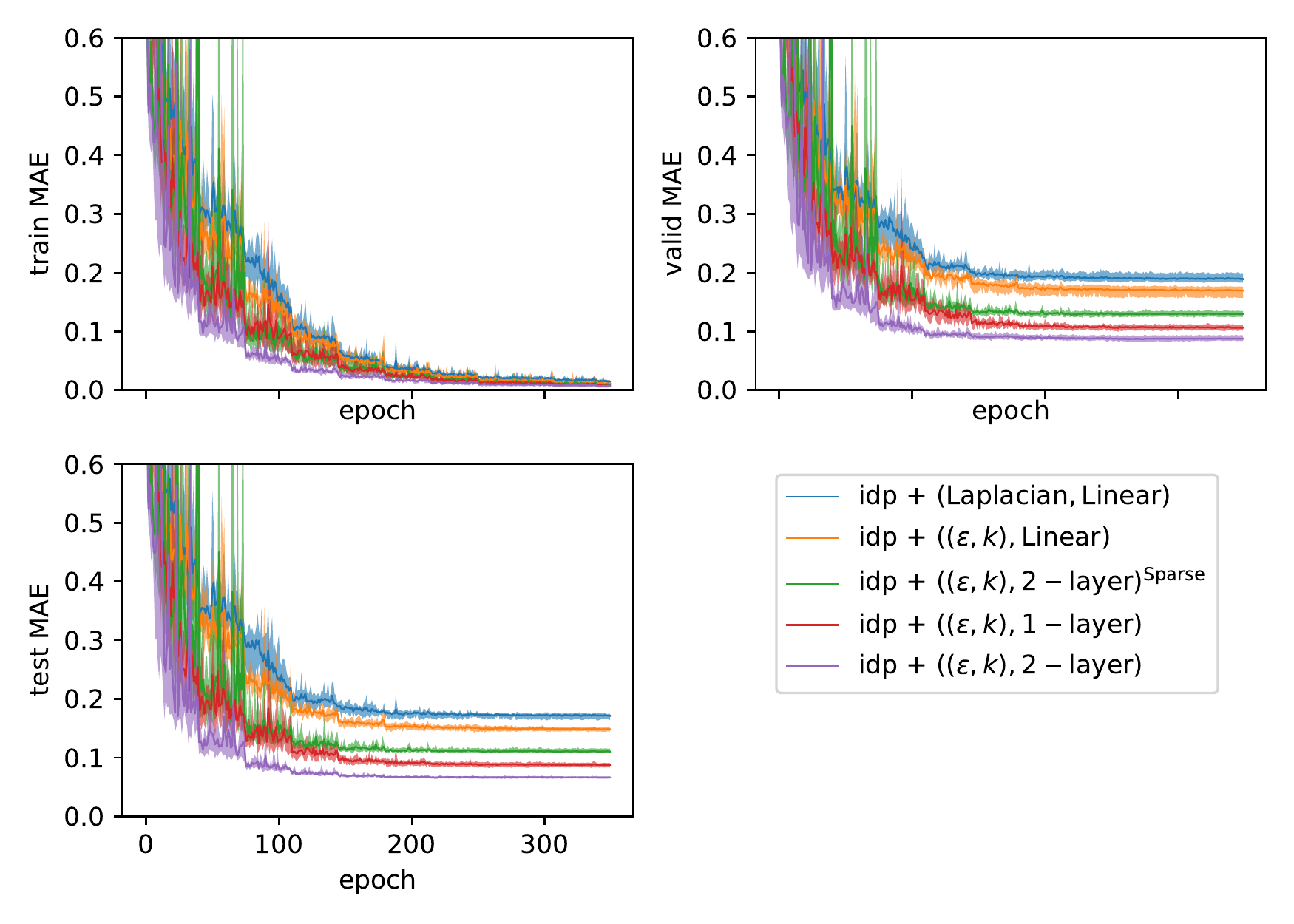}
		\vspace{-10pt}
		\caption{Ablation studies with channel-independent architecture on ZINC.}
		\label{fig:ablation_idp}
		\vspace{-10pt}
	\end{figure*}

	We present the learning curves of training, valid and test sets on ZINC in Fig.~\ref{fig:ablation_shd} and Fig.~\ref{fig:ablation_idp}, which are exactly the same runs as that in Tab.~\ref{tab:ablation}.
	But the curves give more prominent results: on all test cases, channel-independent Parameterized-$(\gD, \gF)$ architecture which assigns each channel with an independent $(\gD, \gF)$ is much more robust than channel-shared Parameterized-$(\gD, \gF)$ architecture in different runs.
	
	\subsection{Complexity Analysis and Computational Efficiency.}
	\label{app:complexity}
	We analyze the time and space complexities of our \method\footnote{Many GNN studies discuss the complexity of their models, 
		but the results vary from each other.
		Here, we follow the widely-adopted one used in \cite{wu2019comprehensive,velickovic2018graph,chiang2019cluster}.}.
	The results are presented in Tab.~\ref{tab:complexity}, where $n$, $m$, $l$, $d$ 
	refer to the number of vertices, edges, layers, and hidden dimensions respectively; 
	$g$ denotes the number of attention heads used in multi-head GAT or the number of aggregators used in PNA and ExpC, 
	and $k=|\gG|$ in \method.
	Generally, $k \ll d$, hence, the additional computations of both channel-shared and channel independent \method are minor compared with GCN or GIN.
	Also, in the channel-independent architecture, each individual channel's computation is fully independent and can be parallelized.
	\begin{table}[h]
		\centering
		\caption{Time and space complexities of existing methods and ours.}
		\label{tab:complexity}
		\vspace{5pt}
			\begin{tabular}{l|c|c}
				\toprule
				Method &Time Complexity &Memory Complexity \\
				\midrule
				GCN~\citep{kipf2017semi} &$O(l(md+nd^2))$ &$O(m+l(nd+d^2))$\\
				GIN~\citep{xu2018how} &$O(l(md+nd^2))$ &$O(m+l(nd+d^2))$\\
				GAT~\citep{velickovic2018graph} &$O(lg(md+nd^2))$ &$O(m+lg(nd+d^2))$\\
				PNA~\citep{corso2020pna} &$O(lg(md+nd^2))$ &$O(m+lg(nd+d^2))$\\
				ExpC~\citep{yang2022expc} &$O(lg(md+nd^2))$ &$O(m+lg(nd+d^2))$\\
				\midrule
				shd-\method &$O(mk+l(md+nd^2))$ &$O(km+l(nd+d^2))$\\
				idp-\method &$O(mkd+l(md+nd^2))$ &$O(k(m+d)+l(nd+d^2))$\\
				\bottomrule
			\end{tabular}
		\vspace{-10pt}
	\end{table}
	
	We tested the practical running time on a shared computing cluster environment running on Nvidia A100 (40GiB) GPU server.
	All test codes are built upon Deep Graph Library (DGL).
	Tab.~\ref{tab:zinc_time_stat} presents the running time on ZINC dataset which involves small molecular graphs.
	And we test both dense graph matrix representations and 1-hop sparse graph matrix representations.
	Tab.~\ref{tab:rdtb_time_stat} presents the running time on RDT-B dataset, which involves large social network graphs.
	And we only test 1-hop sparse graph matrix representations.
	The running time results show that all our four test cases on ZINC and all our two test cases on RDT-B have minor differences in computational efficiency.
	And their running time is analogical to GIN.
	Channel-shared architectures almost have the same efficiency compared with channel-independent architecture on both \method and \method$^{\textrm{1-hop}}$.
	On both architectures, \method takes slightly longer time than \method$^{\textrm{1-hop}}$ due to processing densely vertex connections.
	But the differences are also very small.
	In conclusion, the best performing test case idp-\method, which is used in baseline comparisons in Sec.~\ref{sec:results}, has similar training and evaluation speed compared with GIN.
	\begin{table}[h]
		\centering
		\caption{Training and evaluation time on ZINC. mean $\pm$ std over $50$ epochs (seconds).}
		\label{tab:zinc_time_stat}
		\vspace{5pt}
			\begin{tabular}{ll|cccc}
				\toprule
				\multicolumn{2}{c}{Model} &Training (Train set) &Eval (Train set) &Eval (Val set) &Eval (Test set) \\
				\midrule
				\multicolumn{2}{c|}{GIN} &3.016 $\pm$ 0.198 &1.459 $\pm$ 0.083 &0.531 $\pm$ 0.081 &0.533 $\pm$ 0.077 \\
				\midrule
				\multirow{2}{*}{shd-} &\method$^{\textrm{1-hop}}$ &3.539 $\pm$ 0.218 &1.503 $\pm$ 0.078 &0.426 $\pm$ 0.064 &0.430 $\pm$ 0.056\\
				&\method &4.024 $\pm$ 0.211 &1.769 $\pm$ 0.115 &0.588 $\pm$ 0.086 &0.583 $\pm$ 0.078\\
				\midrule
				\multirow{2}{*}{idp-} &\method$^{\textrm{1-hop}}$ &3.638 $\pm$ 0.184 &1.599 $\pm$ 0.089 &0.572 $\pm$ 0.076 &0.574 $\pm$ 0.089 \\
				&\method &4.050 $\pm$ 0.174 &1.744 $\pm$ 0.086 &0.622 $\pm$ 0.091 &0.613 $\pm$ 0.071 \\
				\bottomrule
			\end{tabular}
		\vspace{-10pt}
	\end{table}
	
	\begin{table}[h]
		\centering
		\caption{Training and evaluation time on RDT-B. mean $\pm$ std over $50$ epochs (seconds).}
		\label{tab:rdtb_time_stat}
		\vspace{5pt}
			\begin{tabular}{l|ccc}
				\toprule
				Model &Training (Train set) &Eval (Train set) &Eval (Val set) \\
				\midrule
				GIN &0.792 $\pm$ 0.009 &0.335 $\pm$ 0.004 &0.040 $\pm$ 0.002 \\
				\midrule
				shd-\method$^{\textrm{1-hop}}$ &0.723 $\pm$ 0.005 &0.301 $\pm$ 0.003 &0.035 $\pm$ 0.001 \\
				\midrule
				idp-\method$^{\textrm{1-hop}}$ &1.035 $\pm$ 0.006 &0.407 $\pm$ 0.006 &0.049 $\pm$ 0.002 \\
				\bottomrule
			\end{tabular}
		\vspace{-10pt}
	\end{table}
	
	\subsection{Ablation Settings}
	\label{appe:ablation_setting}
	
	\begin{table}[h]
		\centering
		\caption{Ablation study settings on ZINC.}
		\label{tab:ablation_settings}
		\vspace{5pt}
			\begin{tabular}{c|ccccc}
				\toprule
				Hyperparameter &$(\textrm{Lap}, \textrm{Lin})$ &$((\eps, k), \textrm{Lin})$ &$((\eps, k), \textrm{2L})^{\textrm{sps}}$ &$((\eps, k), \textrm{1L})$ &$((\eps, k), \textrm{2L})$\\
				\midrule
				Hidden Dim. &\multicolumn{5}{c}{160}\\
				Num. Layers &\multicolumn{5}{c}{6}\\
				Drop. Rate &\multicolumn{5}{c}{0}\\
				Readout &\multicolumn{5}{c}{Mean}\\
				\midrule
				Batch Size &\multicolumn{5}{c}{64}\\
				Initial LR &\multicolumn{5}{c}{0.001}\\
				LR Dec. Steps &\multicolumn{5}{c}{35}\\
				LR Dec. Rate &\multicolumn{5}{c}{0.6}\\
				\# Warm. Steps &\multicolumn{5}{c}{5}\\
				Weight Dec. &\multicolumn{5}{c}{5e-5}\\
				\midrule
				$\gG$ &Dense &Dense &Sparse &Dense &Dense\\
				$\sigma$ in $f_{\vtheta}$ &Linear &Linear &GELU &GELU &GELU\\
				\midrule
				$\{(\eps, k)\}$ &
				\makecell{(-0.5, 1),\\(-0.5, 2),\\(-0.5, 3),\\(-0.5, 4),\\(-0.5, 5)} &
				\makecell{(-0.1, 3),\\(-0.2, 3),\\(-0.3, 4),\\(-0.4, 4),\\(-0.5, 4)} &
				\makecell{(-0.1, 4),\\(-0.2, 4),\\(-0.3, 4),\\(-0.4, 4),\\(-0.5, 4)} &
				\makecell{(-0.1, 4),\\(-0.2, 4),\\(-0.3, 4),\\(-0.4, 4),\\(-0.5, 4)} &
				\makecell{(-0.1, 4),\\(-0.2, 4),\\(-0.3, 4),\\(-0.4, 4),\\(-0.5, 4)}\\
				\bottomrule
			\end{tabular}
		\vspace{-10pt}
	\end{table}
	Tab.~\ref{tab:ablation_settings} summaries the model setting details in our ablation studies.
	To ensure a fair comparison, all five cases share the same hyperparameter setting, 
	which is also the same as that used in baseline comparisons in Tab.~\ref{tab:zinc_ogbg_settings}.
	Also, they apply the same number of input graph matrix representations, i.e. $|\gG|=|\{(\eps, k)\}|$.
	In $((\eps, k), \textrm{Lin})$ case, the applied $\{(\eps, k)\}$ is 
	slightly different with that in the case $((\eps, k), \textrm{2L})$ because in our test, 
	$((\eps, k), \textrm{Lin})$ did not get the best performance 
	when sharing the same $\{(\eps, k)\}$ with $((\eps, k), \textrm{2L})$.
	Based on the above settings, all five implementation cases share the same amount of learnable coefficients, $499681$.
	
\end{document}